\documentclass[acmtog,nonacm]{acmart}

\setcopyright{none}
\renewcommand\footnotetextcopyrightpermission[1]{} 
\acmDOI{}
\acmISBN{}
\acmConference{}{}{}
\acmBooktitle{}

\usepackage{makecell}
\usepackage{pifont}
\usepackage{booktabs} 

\usepackage[ruled]{algorithm2e} 

\SetAlFnt{\small}
\SetAlCapFnt{\small}
\SetAlCapNameFnt{\small}
\SetAlCapHSkip{0pt}
\definecolor{linkpurple}{RGB}{180,0,180}

\usepackage{color}
\usepackage{graphicx}
\usepackage{caption}
\usepackage{wrapfig}
\usepackage{multirow}
\usepackage{multicol}
\usepackage{enumitem}
\usepackage{xspace}
\usepackage{amsmath}
\usepackage{microtype}
\usepackage{tabularx}
\usepackage{booktabs}
\usepackage{pgfplots}
\usepackage[most]{tcolorbox}
\usepackage{fancyvrb}
\usepackage{placeins}  
\usepackage{xr}
\definecolor{title_purple}{rgb}{0.65,0.1,0.65}

\definecolor{green}{rgb}{0, 0.5, 0}
\definecolor{orange}{rgb}{0.8, 0.6, 0.2}
\definecolor{red}{rgb}{1.0, 0.0, 0.0}
\definecolor{teal}{rgb}{0.0, 0.4, 0.4}
\definecolor{purple}{rgb}{0.65,0.0,0.65}
\definecolor{saffron}{rgb}{0.95,0.75,0.2}
\definecolor{turquoise}{rgb}{0.0,0.5,0.5}
\definecolor{black}{rgb}{0.0, 0.0, 0.0}
\definecolor{gray}{rgb}{0.5, 0.5, 0.5}

\definecolor{darkpink}{rgb}{0.561, 0.282, 0.427}
\definecolor{darkturquoise}{rgb}{0., 0.81, 0.822}

\makeatletter
\DeclareRobustCommand\onedot{\futurelet\@let@token\@onedot}

\makeatother

\makeatletter
\def\blfootnote{\xdef\@thefnmark{}\@footnotetext}
\makeatother

\providecommand{\ExLoRA}{\textsc{ExLoRA}\xspace}
\providecommand{\AnaLoRA}{AnaLoRA\xspace}
\providecommand{\CSV}{\textsc{CSV}\xspace}              

\providecommand{\SIA}{\textsc{SIA}\xspace}
\providecommand{\flux}{\textsc{Flux.2}\xspace}

\providecommand{\Ecal}{\mathcal{E}}

\providecommand{\Scal}{\mathcal{S}}
\usepackage{color}
\definecolor{green}{rgb}{0, 0.5, 0}
\definecolor{deepred}{rgb}{0.9, 0.15, 0.15}
\definecolor{orange}{rgb}{0.8, 0.6, 0.2}
\definecolor{red}{rgb}{1.0, 0.0, 0.0}
\definecolor{teal}{rgb}{0.0, 0.4, 0.4}
\definecolor{purple}{rgb}{0.65,0,0.65}
\definecolor{saffron}{rgb}{0.75,0.05,0.05}
\definecolor{turquoise}{rgb}{0.0,0.30,0.15}
\definecolor{black}{rgb}{0.0, 0.0, 0.0}
\definecolor{gray}{rgb}{0.5, 0.5, 0.5}

\begin{document}

\title{\textcolor{linkpurple}{SIA}: 
\textcolor{linkpurple}{S}elective 
\textcolor{linkpurple}{I}mage 
\textcolor{linkpurple}{A}nalogy} 

\author{Amirhossein Alimohammadi}
\authornote{Both authors contributed equally to this research.}
\affiliation{%
 \institution{Simon Fraser University}
 \city{Burnaby}
 \country{Canada}
}
\author{Kian Izadpanah}
\authornotemark[1]
\affiliation{%
 \institution{University of Cyprus and CYENS Centre of Excellence}
 \city{Nicosia}
 \country{Cyprus}
}
\affiliation{%
 \institution{Simon Fraser University}
 \city{Burnaby}
 \country{Canada}
}
\author{Bardiya Kariminia}
\affiliation{%
 \institution{Shahid Beheshti University}
 \city{Tehran}
 \country{Iran}
}
\author{Yiorgos Chrysanthou}
\affiliation{%
 \institution{University of Cyprus}
 \city{Nicosia}
 \country{Cyprus}
}
\affiliation{
 \institution{CYENS Centre of Excellence}
 \city{Nicosia}
 \country{Cyprus}
}
\author{Ali Mahdavi-Amiri}
\affiliation{%
 \institution{Simon Fraser University}
 \city{Burnaby}
 \country{Canada}
}

\begin{abstract}
Image analogy transfers a demonstrated transformation from a before/after pair $(A,A')$ to a new query image $B$. Existing methods assume all changes in $(A,A')$ should be fully transferred, limiting flexibility when users want only a subset of the demonstrated edits. We introduce \emph{Selective Image Analogy} (\SIA{}), a new setting where a single demonstration pair encodes multiple edits, and the user selects at inference time which edits to apply to $B$. To support this, we curate a multi-edit analogy dataset with full-edit and partial-edit targets. We then train a model that first learns to transfer the complete demonstrated transformation, and then gains selective control through a mask-aligned suppression mechanism: a token-gated LoRA that restricts adapter updates to suppressed edit tokens during denoising, paired with a concept suppression vector that weakens the selected tokens in the text embeddings before denoising. Together, these components allow \SIA{} to suppress unwanted edits that remain visible in $A'$, while preserving the user-selected ones, all without requiring a new demonstration pair for each edit combination. We perform extensive quantitative and qualitative evaluations along with user studies to demonstrate the efficacy of our approach.

\textbf{Project Page:} 
\href{https://selective-image-analogy.github.io/}
{\textcolor{linkpurple}{selective-image-analogy.github.io}}
\end{abstract}
\begin{teaserfigure}
\centering
  \includegraphics[width=0.85\textwidth]{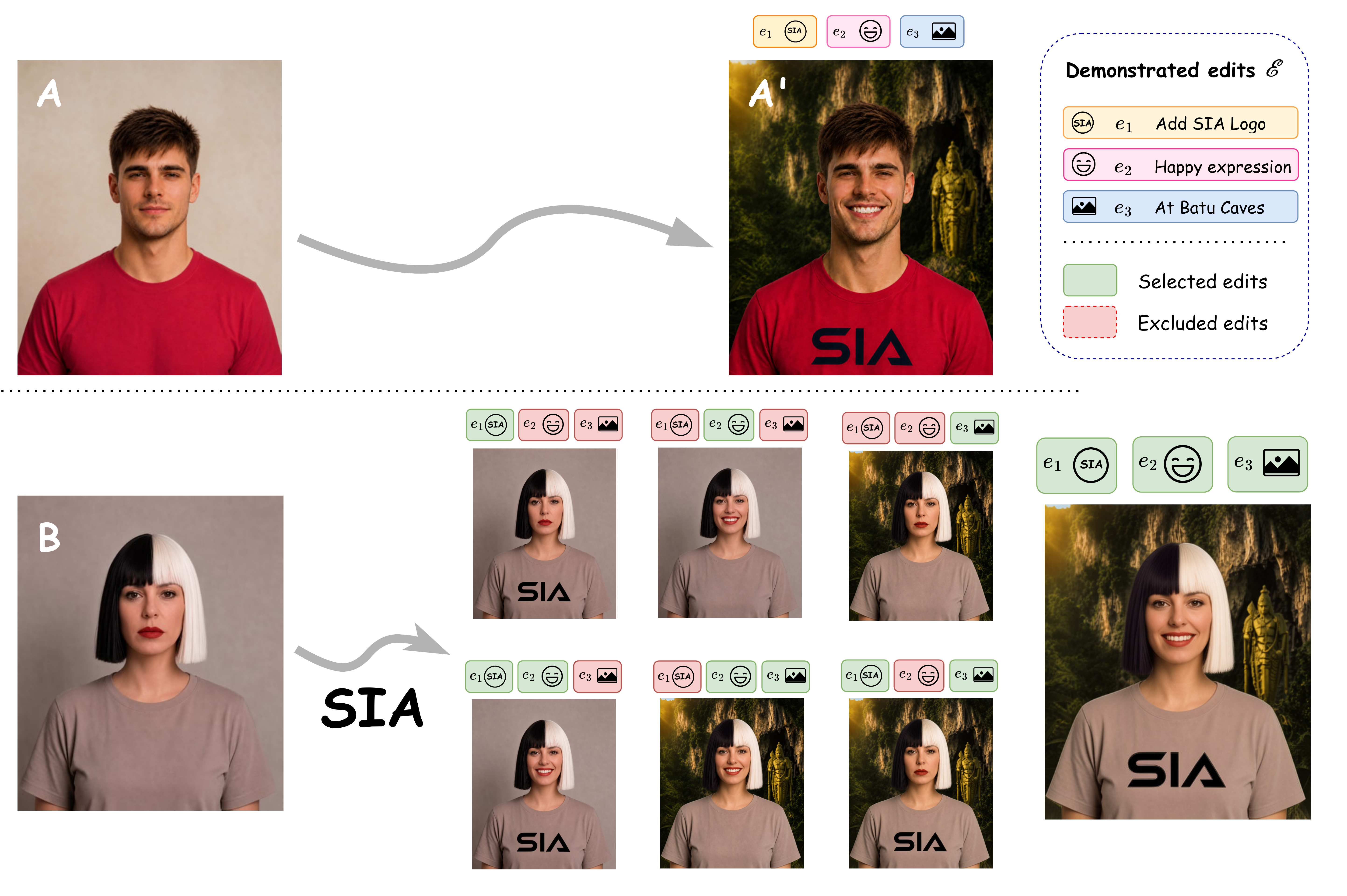}
  \caption{ Teaser of \SIA{}. A single reference pair $(A,A')$ may contain multiple edits at once. In this example, the changes include adding a logo, changing the expression, and replacing the background. Given a new query image $B$, \SIA{} lets the user choose which edits to transfer. The same demonstration can produce different outputs, ranging from single-edit results to combinations of selected edits, while leaving the remaining edits out. 
}
  \label{fig:teaser}
\end{teaserfigure}

\maketitle
\thispagestyle{empty} 

\section{Introduction}
\label{ref:intro}
Image editing plays a crucial role across a wide range of applications, including photography, film production, advertising, medical imaging, and content creation, where precise control over visual details is essential. In recent years, text-based image editing has emerged as a powerful and intuitive approach, enabling users to modify images through natural language instructions~\cite{couairon2022diffedit,zhang2023magicbrush}. While effective, relying solely on text is often insufficient, as language may struggle to capture fine-grained visual nuances, spatial relationships, or stylistic intent ~\cite{li2023gligen,mou2024t2iadapter}. As a result, complementary techniques have been developed to bridge this gap, such as image composition methods that combine multiple visual elements~\cite{yang2023paintbyexample,song2023objectstitch}, sketch-based editing that provides explicit structural guidance~\cite{mikaeili2023sked}, and frameworks like ControlNet~\cite{zhang2023controlnet} that incorporate additional conditioning signals. Together, these approaches enhance controllability and enable more accurate and expressive image editing workflows.

An alternative and powerful editing approach is image analogy, which enables the transfer of transformations from one example to another in a structured way~\cite{hertzmann2001imageanalogies,vsubrtova2023diffusion, chen2025edittransfer}. In this setting, a pair of images $A$ and $A'$ represents a specific edit or transformation—such as a change in style, texture, or geometry. The goal is then to apply the same transformation to a new input image $B$, producing a corresponding output $B'$ that reflects the edits observed between $A$ and $A'$. In other words, the system learns the mapping from $A$ to $A'$ and generalizes it to $B$, allowing consistent and controllable edits without requiring explicit textual or manual specification. This makes image analogy particularly useful for tasks where visual changes are easier to demonstrate than to describe. A text prompt or reference image specifies the category of an edit, while a before/after pair captures its exact visual form—including style, shape, size, placement, magnitude, and subject-specific interactions. These relative properties are difficult to convey through text or a single reference image. For example, in Fig.~\ref{fig:teaser}, text can specify adding an ``SIA'' logo but not its exact font; the demonstration pair directly provides this visual detail, enabling its transfer to the query subject.

However, prior image analogy methods typically assume that all changes observed between $A$ and $A'$ should be fully transferred to the target image $B$. This all-or-nothing behavior limits their flexibility in practical scenarios, where users may only want to apply a subset of the edits. For example, one might wish to transfer attributes such as a smile or the presence of sunglasses from $A'$ while preserving the original pose, lighting, or background of $B$. By entangling multiple transformations into a single mapping, these approaches make it difficult to selectively control which aspects are propagated, thereby restricting their applicability in more nuanced and user-driven editing tasks.

To address this limitation, we introduce a new problem setting termed \emph{Selective Image Analogy}, where the transformation from $A$ to $A'$ may involve multiple edits, but only a subset of these edits is transferred to a target image $B$ (see Fig.~\ref{fig:teaser}). Crucially, the selection of which edits to apply is guided by text, allowing users to specify their intent at a high level while still leveraging the rich visual cues embedded in the example pair. This formulation combines the strengths of text-based editing—flexibility and semantic control—with the expressiveness of image analogy, enabling more precise and user-driven manipulation of visual content.

To solve this problem, we design \SIA{} so that the same demonstration pair $(A,A')$ can support different user-selected outputs. The model first learns the standard analogy behavior, transferring the full transformation shown between $A$ and $A'$ to a new query image $B$. We then make this behavior selective through an edit-token suppression mask, which marks the edits the user does not want to transfer. This mask controls suppression at two aligned points: during denoising,  it applies a token-gated LoRA only to the suppressed edit tokens; before denoising, it weakens the corresponding text embeddings along edit directions estimated from positive and negative prompt pairs. This allows \SIA{} to suppress unwanted edits that remain visible in $A'$, while preserving the edits selected by the user without requiring a new demonstration pair for each edit combination.
We demonstrate the efficacy and generalizability of our method through extensive quantitative and qualitative evaluations, as well as a series of ablation studies that justify our design choices, all while maintaining a lightweight and training-efficient framework that requires no per-query optimization or new demonstration pairs. Therefore, our contributions are as follows:

\begin{itemize}
    \item We introduce \emph{Selective Image Analogy}, a new image analogy setting that, to our knowledge, is the first to allow a multi-edit before/after demonstration to be selectively transferred to a new image.

    \item We introduce a mask-aligned suppression mechanism that localizes suppression to user-selected edit phrases, enabling unwanted edits to be blocked even when they remain visible in the demonstration.

     \item We also curate multi-edit training examples with full-edit and partial-edit targets, allowing the model to learn both complete and selective transfer.
\end{itemize}
\section{Related Works}

\subsection{Image editing.}
Diffusion-based image editing was popularized by SDEdit~\cite{meng2021sdedit}, which edits an input image by adding noise and denoising it under a target condition. Later, instruction-based editors made this process more accessible by allowing users to describe edits with natural-language commands~\cite{brooks2023instructpix2pix,kawar2023imagic,mokady2023nulltext,brack2024ledits}. 

Since text alone is often not enough for precise control, recent works incorporate additional signals such as structural guidance with ControlNet~\cite{zhang2023controlnet}, reference-based conditioning~\cite{ye2023ipadapter}, and attention-sharing mechanisms~\cite{cao2023masactrl, alimohammadi2025cora, mikaeili2026untwisting, xu2024infedit}. More recently, transformer-based diffusion and flow models have shown strong versatility across local and global editing tasks~\cite{labs2025fluxkontext,qwen2025imageedit}. Our work builds on this progress, but moves toward selective visual analogies, where the edit is given by a before/after demonstration and only the user-selected parts of that demonstration are transferred.

\subsection{Image analogy.}
In image analogy, the user demonstrates an edit with a before/after example, and the edit is transferred to another image, which shares similarity with the example pair. Classical Image Analogies~\cite{hertzmann2001imageanalogies} introduced the general formulation of transferring the edit through example-based local visual matching. Deep Image Analogy~\cite{Liao2017DeepImageAnalogy} extended this idea to semantically related image pairs by matching deep features, enabling visual attribute transfer across large appearance gaps, while Structural Analogy~\cite{Benaim2021StructuralAnalogy} learns a generative mapping between the structures of two images from a single pair. With the advent of stronger generative models such as diffusion models, image analogy was revisited using noise inversion techniques guided by the CLIP embedding space~\cite{vsubrtova2023diffusion}.
Alternative approaches leveraged visual prompting and in-context learning, where example pairs are placed together with the query image as visual context~\cite{bar2022visualprompting,wang2023painter,sun2023imagebrush,gu2024analogist}. The intuition is that, when these images are given jointly to a diffusion transformer, the model can compare them through attention and infer the transformation to apply, without optimizing a separate model for each test example.

Building on this direction, Edit Transfer~\cite{chen2025edittransfer} learns visual relations from source-target editing examples and applies them to new query images. LoRWeB~\cite{manor2026lorweb} improves generalization by dynamically composing a basis of LoRA modules for different analogy types. VIRAL~\cite{li2026viral} further studies visual in-context reasoning with diffusion transformers using role-aware multi-image conditioning and expert mixing. These methods show that visual demonstrations can capture transformations that are difficult to express with language alone, including spatial, non-rigid, and compositional edits. However, they generally transfer the demonstrated transformation as a whole. In contrast, our setting considers demonstrations where $A \rightarrow A'$ contains multiple edits, and the user may want to transfer only a selected subset while keeping the same demonstration fixed.

\subsection{Controllable image editing.}
Our work broadly falls within the domain of controllable image editing, where users are given more direct control over which visual changes are applied and how strongly they appear. Concept Sliders~\cite{gandikota2024conceptsliders} learns LoRA sliders for visual concepts, allowing continuous control over semantics. TARA~\cite{peng2026tara} uses token masks to reduce interference between LoRA modules in multi-concept personalization. SliderEdit~\cite{zarei2025slideredit} similarly restricts LoRA updates to selected instruction tokens, enabling fine-grained control over different parts of a multi-instruction edit.

Another line of work controls generation and editing through meaningful directions in learned representation spaces. Early methods study such directions in GAN latent spaces~\cite{shen2020interfacegan} or use CLIP guidance to manipulate StyleGAN representations~\cite{patashnik2021styleclip}. More recent diffusion and flow-based methods apply similar ideas to text embeddings and intermediate activations. Continuous Subject-Specific Attribute Control~\cite{baumann2025continuous} estimates CLIP directions from contrastive prompts. SHIFT~\cite{konovalova2026shift} builds steering vectors from paired positive and neutral prompts and applies them to intermediate activations in DiT-based flow models, while Diffusion Sliders~\cite{ekin2026steering} derives text-encoder directions from contrastive prompt pairs for controllable steering.

These works control visual changes through token-level mechanisms, adapters, and representation-space directions, and our proposed method shares this spirit. However, applying these techniques to selective image analogy is non-trivial: the edit to suppress is not absent from the input---it remains fully visible in $A'$---so the model cannot simply ignore it. Text removal alone is insufficient, since the visual demonstration actively reintroduces the suppressed edit signal. This helps suppression operate at two aligned stages: weakening the text-conditioning signal before denoising and gating the adapter updates during denoising, both localized to the specific edit tokens. Our work is therefore the first to make a single visual analogy demonstration selectable, allowing a subset of edits to be transferred from the same fixed before/after pair.

\section{Preliminaries}
\label{sec:preliminaries}
\noindent\textbf{Flow Models.}
Flow-based generative models~\cite{tong2024conditionalflowmatching, lipman2023flowmatching,liu2023rectifiedflow} learn a velocity field that transports a noisy latent sample toward data distribution. Let $x_0$ denote latent of clean data and $x_1 \sim \mathcal{N}(0,I)$ denote a noise sample. A noised latent at time $t$ is obtained by interpolation,
\begin{equation}
    z_t = (1-t)x_0 + t x_1,
\end{equation}
and the model is trained to predict the velocity from $x_0$ to $x_1$. In the \emph{conditional} setting, the velocity network also receives a text prompt $c$ and visual conditioning input $y$. The training objective is
\begin{equation}
    \mathcal{L}_{\mathrm{FM}}
    =
    \mathbb{E}_{t,x_0,x_1,y,c}
    \left[
    \left\|
    v_{\theta}(z_t,t,y,c) - (x_1-x_0)
    \right\|_2^2
    \right].
\end{equation}

\noindent\textbf{\flux-Klein{} backbone.}
\flux-Klein{}~\cite{blackforest2025flux2} follows a multimodal diffusion transformer design in which image and text tokens are processed jointly. Images are first encoded into latent tokens by a VAE, while the text prompt is encoded by a Qwen3 text encoder~\cite{qwen3}. The transformer $f_\theta$ then denoises the target latent conditioned on both the visual stream and the text stream. This architecture is well-suited for our setting because selective image analogy requires the model to reason jointly over the visual demonstration $(A,A')$, the target image $B$, and the text tokens that name the edits.

\noindent\textbf{Low-rank adaptation.}
LoRA~\cite{hu2022lora} adapts a large pre-trained model efficiently without updating all of its weights. Given a frozen linear layer with weight matrix $W_0 \in \mathbb{R}^{m \times n}$, LoRA represents the trainable update as a low-rank decomposition $\Delta W = BA$,
where $B \in \mathbb{R}^{m \times r}$, $A \in \mathbb{R}^{r \times n}$, and $r \ll \min(m,n)$. The adapted layer is then written as:
\begin{equation}
    W = W_0 + \frac{\alpha}{r}BA,
\end{equation}
where $\alpha$ controls the scale of the update. Since only $A$ and $B$ are trained, LoRA preserves the pre-trained model while allowing efficient task-specific adaptation.

\section{Method}
\label{sec:method}

\begin{figure*}[t]
    \centering
    \includegraphics[width=0.9\textwidth]{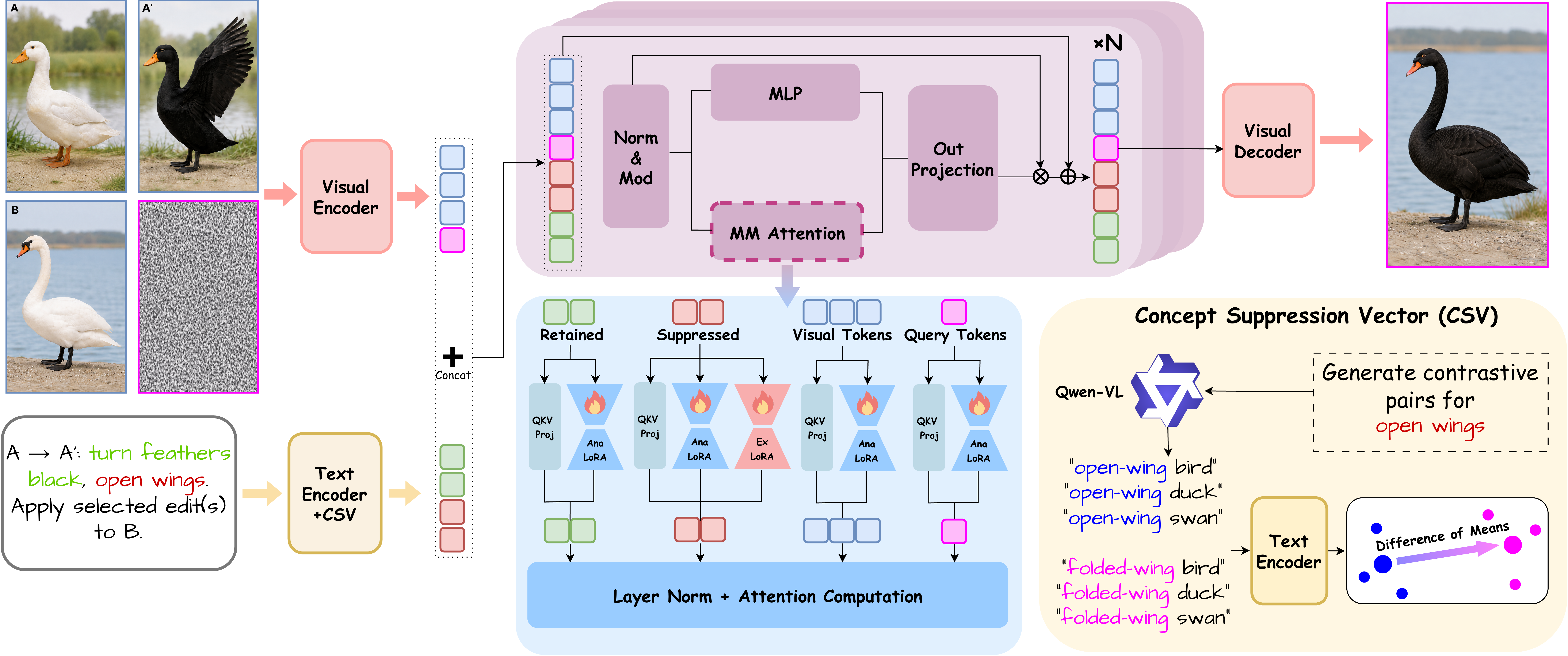}
    \caption{
        \textbf{Overview of our selective image analogy pipeline}. Given a demonstration pair $(A,A')$ containing multiple edits and a query image $B$, the model transfers only the user-selected edits to produce $B'$. In this example, \textcolor{green!60!black}{green textual tokens} indicate retained edits that should be applied to $B$, while \textcolor{red!75!black}{red textual tokens} indicate suppressed edits that should not be transferred, even though they are present in $A'$. The visual analogy guides the overall edit transfer, while the colored textual tokens specify which edits are retained or suppressed using selective token suppression and the Concept Suppression Vector (CSV).
    }
\label{fig:method}
\end{figure*}

We propose \SIA{}, a framework for \emph{Selective Image Analogy} using a visual before/after demonstration. Given a source image $A$, its edited version $A'$, and a target image $B$, our goal is to transfer only the user-selected edits from the change $A \rightarrow A'$ onto $B$ to produce $B'$, as illustrated in Figure~\ref{fig:teaser}.
The transformation from $A$ to $A'$ may contain multiple edits $\mathcal{E}=\{e_1,e_2,\ldots,e_N\}$, such as a pose change, an expression change, a style transfer, or an object addition/removal. However, the user may specify a subset $\mathcal{R}\subseteq\mathcal{E}$ to retain, leaving the suppressed edits $\mathcal{S}=\mathcal{E}\setminus\mathcal{R}$ to be blocked in the output $B'_{\mathcal{R}}$. 

Unlike standard image analogy~\cite{hertzmann2001imageanalogies}, which transfers the full demonstrated transformation, \SIA{} keeps the same before/after pair fixed while allowing selective control over which edits are applied. This setting differs from ordinary text-guided editing because the model also receives the visual pair $(A,A')$ as conditioning: since $A'$ already contains all demonstrated edits, simply removing an edit from the prompt is insufficient---the model can still infer and transfer it from $A'$. Rather than constructing a separate demonstration pair for each possible subset (which would be impractical), we use one full demonstration pair and control which edits are transferred at inference time via a token-level suppression mask that marks phrases corresponding to suppressed edits; \SIA{} first learns to transfer all demonstrated edits and then applies this mask to selectively block them, as described below.

\subsection{Full Transfer Model}
\label{subsec:default_analogy}

Before introducing selective control, we first build the dataset and train the model to perform full visual analogy, both of which we describe in the following.

\noindent\textbf{Dataset.} Since no existing dataset supports selective multi-edit analogy, we construct one tailored to our task. Images are organized by \emph{edit recipes}, each specifying a set of edits $\mathcal{E}=\{e_1,\ldots,e_N\}$ with $N\in\{2,3,4\}$, where each group shows the same subject with different combinations of edits applied. For each subject in a recipe, we store the original image $I^{\emptyset}$, the fully edited image $I^{\mathcal{E}}$, and partial images $I^{\mathcal{R}}$ for all retained subsets $\mathcal{R}\subseteq\mathcal{E}$. Training tuples are formed by pairing two subjects $a$ and $b$ from the same recipe, setting $A=I_a^{\emptyset}$, $A'=I_a^{\mathcal{E}}$, $B=I_b^{\emptyset}$, with targets $B'_{\mathcal{E}}=I_b^{\mathcal{E}}$ and $B'_{\mathcal{R}}=I_b^{\mathcal{R}}$ supervising full and selective transfer respectively. 

The two-edit group covers pairwise combinations of object addition, style change, and background change, built by augmenting a subset of RelationAdapter~\cite{gong2025relationadapter} with a second edit synthesized via \flux-Klein{}. The three- and four-edit groups are constructed by sequentially applying edits to source images, e.g.\ $I^{\emptyset} \rightarrow I^{\{e_1\}} \rightarrow I^{\{e_1,e_2\}} \rightarrow I^{\mathcal{E}}$, covering combinations of object additions, style changes, and background changes. All samples are filtered with a vision-language model to verify that desired edits are present and excluded edits are absent. In total, the dataset contains 1606 training samples drawn from 238 edit settings and 239 test cases, covering 2–4 simultaneous edits over both creature and non-creature subjects. Further details are provided in the supplementary material (Sec.~\ref{supp:dataset}).

\noindent\textbf{\AnaLoRA.} Given a demonstration pair $(A,A')$ and a query image $B$, the model learns to transfer all demonstrated edits $\mathcal{E}=\{e_1,\ldots,e_N\}$ and apply them to $B$, producing the full-edit output $B'_{\mathcal{E}}$. We utilize \flux-Klein{} and apply a lightweight LoRA adapter, which we call \emph{\AnaLoRA{}}, trained to internalize the complete demonstrated transformation before any selective suppression is applied.

We adopt the in-context editing formulation of~\cite{chen2025edittransfer}, based on a four-image tuple $(A,A',B,B')$ representation. The latents of $A$, $A'$, and $B$ are used as visual conditioning inputs, while $B'$ denotes the target image to be generated. Let $z_A$, $z_{A'}$, $z_B$, and $z_{B'_{\mathcal{E}}}$ denote the latent tokens of the source image, edited source image, query image, and full-edit target, respectively. We pack the conditioning latents as $z_{\mathrm{cond}} = [\, z_A;\, z_{A'};\, z_B \,]$, where $[\,\cdot;\cdot\,]$ denotes concatenation along the token dimension. The visual conditioning is paired with a structured analogy prompt that names all demonstrated edits: 

\begin{equation}
\begin{aligned}
    \mathcal{P}(\mathcal{E}) =\;&
    \texttt{"Image 1 is the original image."} \\
    &\texttt{"Image 2 is the edited version of Image 1."} \\
    &\texttt{"The edits are: } e_1, e_2, \dots, e_N \texttt{."} \\
    &\texttt{"Apply the same edits to Image 3."}
\end{aligned}
\end{equation}
This prompt is encoded by the text encoder as $\phi_{\mathrm{text}}(\mathcal{P}(\mathcal{E}))$ and injected into the MM-DiT text stream. The denoising transformer receives both the visual demonstration $(A,A')$ and a text description of the edits to transfer.
After this stage, the model has learned the default full-transfer behavior: given the complete demonstration, it transfers all edits in $\mathcal{E}$ to the query image. We then make this behavior controllable, so the same demonstration pair $(A,A')$ can be used while selected edits are excluded.

\subsection{Concept Suppression}
\label{subsec:mask_aligned_suppression}
Since suppressed edits remain visible in $A'$, simply removing them from the prompt is insufficient---the model can still infer and transfer them from the visual demonstration. Our goal is therefore to suppress only the excluded edits while keeping the demonstration pair fixed.
We address this by conditioning selective transfer on a suppression mask. Given the user-selected retained set $\mathcal{R}\subseteq\mathcal{E}$, we define the suppressed set as $\mathcal{S}=\mathcal{E}\setminus\mathcal{R}$. Let $\tau(e_i)\subseteq \{1, ..., T\}$ denote the set of token positions of the phrase describing edit $e_i$, where $T$ is the text sequence length. We construct a binary token mask $M(\mathcal{S}) \in \{0,1\}^{T}$ with $M_j(\mathcal{S})=1$ if $j \in \bigcup_{e_i \in \mathcal{S}} \tau(e_i)$ and $0$ otherwise. Thus, $M(\mathcal{S})$ marks the edit phrases that should be excluded from transfer, while leaving the unmasked edit phrases available to guide generation.
We use the same mask in the two suppression operations described next. \ExLoRA{} changes how the denoising transformer handles the masked edit tokens during denoising, while \CSV{} weakens the same edit phrases in the prompt embeddings before denoising.

\noindent\textbf{ExLoRA.} To make the suppression mask affect denoising, we introduce \emph{\ExLoRA{}}, a token-gated LoRA~\cite{zarei2025slideredit} that applies its low-rank update only to the text tokens marked by the suppression mask---those corresponding to demonstrated edit phrases that should be excluded from the output.
While \AnaLoRA{} (full edit) remains active for the full-transfer analogy behavior, \ExLoRA{} handles the edit phrases marked by the suppression mask. It does not learn a separate analogy model. Instead, it changes how the denoising transformer handles the masked edit phrases, allowing the model to use the same full demonstration while suppressing only the excluded edits.
Formally, let $h_j^\ell$ denote the hidden representation of text token $j$ at layer $\ell$, and let $W^\ell$ be a frozen linear projection in the text stream. While a standard LoRA update applies the same residual to every token, \ExLoRA{} gates this residual by the suppression mask:
\begin{equation}
    \tilde{h}_j^\ell
    =
    \left(
    W^\ell
    +
    M_j(\mathcal{S})
    \frac{\alpha}{r}
    B_s^\ell A_s^\ell
    \right)
    h_j^\ell,
    \qquad j=1,\ldots,T.
    \label{eq:mask_gated_ExLoRA}
\end{equation}
Here, $A_s^\ell$ and $B_s^\ell$ are the selective LoRA weights, $r$ is the rank, and $\alpha$ is the LoRA scaling factor. When $M_j(\mathcal{S})=0$, token $j$ is processed by the ordinary text stream. When $M_j(\mathcal{S})=1$, the selective LoRA residual is applied, so the model treats the corresponding edit phrase as suppressed.

Suppression is therefore localized to the masked edit phrase. The model does not require a new demonstration pair, a separate prompt for each subset, or a query-specific adapter. Instead, it receives the same full analogy input together with a token mask specifying which edit descriptions should be suppressed.

\noindent\textbf{Concept Suppression Vector.}
While \ExLoRA{} modifies how the denoising transformer responds to masked tokens, the prompt embeddings still encode all edits, including suppressed ones. To remove this residual signal in the text conditioning, we apply a token-local Concept Suppression Vector (\CSV{}) before denoising.

For each edit $e_i$, we estimate an edit direction $d_i$ in the text-encoder embedding space. We use contrastive prompt pairs in which the edit is present or absent. As a result, given positive and negative prompt pairs $\{(p_n^+,p_n^-)\}_{n=1}^{N}$, we encode them with the same text encoder used by \flux-Klein{} and pool the hidden states corresponding to the target edit phrase. Let $u_n^+$ and $u_n^-$ denote the pooled positive and negative embeddings. We compute
\begin{equation}
    d_i =
    \frac{
    \bar{u}_i^+ - \bar{u}_i^-
    }{
    \left\|
    \bar{u}_i^+ - \bar{u}_i^-
    \right\|_2
    },
    \qquad
    \bar{u}_i^+ = \frac{1}{N}\sum_{n=1}^{N}u_n^+,
    \quad
    \bar{u}_i^- = \frac{1}{N}\sum_{n=1}^{N}u_n^- .
    \label{eq:mask_aligned_direction}
\end{equation}
The resulting direction is computed once for each edit phrase and cached.
At inference time, if edit $e_i$ belongs to the suppressed set $\mathcal{S}$, we apply \CSV{} only to the token positions corresponding to that edit phrase. For each token embedding $h_j$ with $j\in \tau(e_i)$, we remove its component along the edit direction and then push it away from the edit:
\begin{equation}
    h_j
    \leftarrow
    h_j
    -
    (h_j^\top d_i)\,d_i
    -
    \lambda_i\, d_i,
    \qquad j\in \tau(e_i),
    \label{eq:mask_aligned_csv}
\end{equation}
where $\lambda_i \ge 0$ is the suppression strength. The projection term removes the edit concept from the token, and $\lambda_i$ sets how far the token is then moved against the edit direction; $\lambda_i = 0$ performs pure removal. Tokens corresponding to retained edits and general analogy instructions are left unchanged.

\CSV{} and \ExLoRA{} operate at different stages using the same mask: \CSV{} weakens suppressed edit tokens in the prompt embeddings before denoising, while \ExLoRA{} modifies how the transformer processes them during denoising, together reducing leakage of suppressed edits into the output.
\begin{table*}[t]
\caption{Quantitative comparison with visual analogy baselines. VLM scores
are reported using Qwen3-VL-32B-Instruct and Gemma 4; results with Gemma 3 as
an additional judge are given in the supplementary material
(Tab.~\ref{tab:quantitative_full}). Ret. Acc. measures retained-edit quality,
Pres. measures preservation of $B$, and Recall measures retained-edit recall.
CLIP-Dir Image is the image-image directional metric, while CLIP-Dir Retained
measures alignment with the retained edit direction.}
\label{tab:quantitative}
\small
\setlength{\tabcolsep}{4pt}
\begin{tabular*}{\textwidth}{@{\extracolsep{\fill}} l ccc ccc cc}
\toprule
& \multicolumn{3}{c}{Qwen3-VL-32B} & \multicolumn{3}{c}{Gemma 4}
& \multicolumn{2}{c}{CLIP} \\
\cmidrule(lr){2-4}\cmidrule(lr){5-7}\cmidrule(lr){8-9}
Method & Ret. Acc.$\uparrow$ & Pres.$\uparrow$ & Recall$\uparrow$
       & Ret. Acc.$\uparrow$ & Pres.$\uparrow$ & Recall$\uparrow$
       & Dir. Image$\uparrow$ & Dir. Ret.$\uparrow$ \\
\midrule
RelationAdapter~\cite{gong2025relationadapter}
  & 6.74 & 9.93 & 0.782 & 6.14 & 9.50 & 0.706 & 0.386 & 0.144 \\
VisualCloze~\cite{li2025visualcloze}
  & 7.24 & 8.62 & 0.856 & 6.35 & 7.53 & 0.743 & 0.464 & 0.152 \\
Edit Transfer~\cite{chen2025edittransfer}
  & 2.31 & \textbf{9.96} & 0.176 & 2.35 & 9.83 & 0.181 & 0.102 & 0.023 \\
LoRWeB~\cite{manor2026lorweb}
  & 6.59 & 9.83 & 0.736 & 5.27 & 9.72 & 0.567 & 0.402 & 0.155 \\
\midrule
\textbf{Ours}
  & \textbf{8.01} & 9.75 & \textbf{0.892}
  & \textbf{9.31} & \textbf{9.89} & \textbf{0.965} & \textbf{0.543} & \textbf{0.171} \\
\bottomrule
\end{tabular*}
\end{table*}

\subsection{Training and Inference}
\label{subsec:training_inference}

We now describe the training procedure and inference process. Training uses both full-transfer and selective-suppression examples. The learned adapters, \AnaLoRA{} and \ExLoRA{}, are optimized with the flow-matching loss, while \CSV{} is computed separately from contrastive text pairs and applied only at inference time.

We use the same visual conditioning latents throughout training,
\[
    z_{\mathrm{cond}} = [\, z_A;\, z_{A'};\, z_B \,],
\]
where $z_A$, $z_{A'}$, and $z_B$ are the latent tokens of the source image, edited source image, and query image, respectively.

In both training cases, the model receives the same visual conditioning input $(A,A',B)$ and the same full analogy prompt $\mathcal{P}(\mathcal{E})$. The difference is whether the suppression mask is inactive or active, and whether the target contains all demonstrated edits or only a retained subset.

\noindent\textbf{Full-transfer step.}
In a full-transfer step, the model receives the demonstration pair $(A,A')$, the query image $B$, and the full prompt $\mathcal{P}(\mathcal{E})$. No edit is suppressed, so the suppression set is empty and the token mask is zero:
\begin{equation}
    \mathcal{S}=\emptyset,
    \qquad
    M(\mathcal{S})=\mathbf{0}.
\end{equation}
The training target is the full-edit image $B'_{\mathcal{E}}$. Given Gaussian noise $\epsilon\sim\mathcal{N}(0,I)$ and flow time $t\in[0,1]$, we form
\begin{equation}
    z_t
    =
    (1-\sigma_t)z_{B'_{\mathcal{E}}}
    +
    \sigma_t\epsilon,
    \qquad
    v^\star_{\mathcal{E}}
    =
    \epsilon - z_{B'_{\mathcal{E}}}.
\end{equation}
The full-transfer prediction is
\begin{equation}
    \hat{v}_{\mathcal{E}}
    =
    f_{\theta,\AnaLoRA}
    \left(
        z_t,
        z_{\mathrm{cond}},
        \phi_{\mathrm{text}}(\mathcal{P}(\mathcal{E})),
        t
    \right),
\end{equation}
and we optimize
\begin{equation}
    \mathcal{L}_{\mathrm{full}}
    =
    \mathbb{E}_{t,\epsilon}
    \left[
    \left\|
    \hat{v}_{\mathcal{E}}
    -
    v^\star_{\mathcal{E}}
    \right\|_2^2
    \right].
    \label{eq:full_transfer_loss}
\end{equation}
This step teaches \AnaLoRA{} the default analogy behavior: given the full demonstration, the model should transfer all demonstrated edits to the query image.

\noindent\textbf{Selective-suppression step.}
In a selective-suppression step, the model still receives the full demonstration $(A,A')$, the query image $B$, and the full prompt $\mathcal{P}(\mathcal{E})$ naming all edits. We sample a retained subset $\mathcal{R}$, construct the suppression mask $M(\mathcal{S})$ over the excluded edit phrases, and train against the partial target $B'_{\mathcal{R}}$, which contains only the retained edits.

For this target, we form
\begin{equation}
    z_t
    =
    (1-\sigma_t)z_{B'_{\mathcal{R}}}
    +
    \sigma_t\epsilon,
    \qquad
    v^\star_{\mathcal{R}}
    =
    \epsilon - z_{B'_{\mathcal{R}}}.
\end{equation}
The selective prediction is
\begin{equation}
    \hat{v}_{\mathcal{R}}
    =
    f_{\theta,\AnaLoRA,\ExLoRA}
    \left(
        z_t,
        z_{\mathrm{cond}},
        \phi_{\mathrm{text}}(\mathcal{P}(\mathcal{E})),
        M(\mathcal{S}),
        t
    \right).
\end{equation}
We optimize the selective flow-matching loss
\begin{equation}
    \mathcal{L}_{\mathrm{sel}}
    =
    \mathbb{E}_{t,\epsilon}
    \left[
    \left\|
    \hat{v}_{\mathcal{R}}
    -
    v^\star_{\mathcal{R}}
    \right\|_2^2
    \right].
    \label{eq:selective_loss}
\end{equation}

This supervision is counterfactual: the input shows and names the full transformation, but the target removes only the masked edits. Thus, the model learns to control which edits are transferred by changing the mask, without needing an input for edit subsets.

\noindent\textbf{Inference.}
At inference time, the user provides one demonstration pair $(A,A')$, one query image $B$, and the edits to retain, $\mathcal{R}$. We define the suppressed edits as $\mathcal{S}=\mathcal{E}\setminus\mathcal{R}$. We keep the full visual conditioning $(A,A',B)$ and the full prompt $\mathcal{P}(\mathcal{E})$ unchanged. We then construct the token mask $M(\mathcal{S})$ over the suppressed edits.

The same mask controls both parts of the selective-conditioning mechanism. It first determines where \CSV{} weakens the prompt embeddings using Eq.~\ref{eq:mask_aligned_csv}. The same token positions then activate \ExLoRA{} inside the text stream through Eq.~\ref{eq:mask_gated_ExLoRA}. The denoising process then generates $B'_{\mathcal{R}}$, which should preserve the retained demonstrated edits while excluding the suppressed ones.
Thus, inference uses one fixed demonstration pair and changes only the suppression mask over the edit tokens and no subset-specific examples or query-specific optimization are required.

\section{Experiments}
\label{sec:experiments}

We evaluate \SIA on multiple demonstrated edits using our Multi-Edit Analogy Dataset. Our goal is to evaluate selective transfer quality, the contribution of \ExLoRA{} and \CSV{} over prompt-based suppression, and robustness across edit types and multi-edit settings. Two user studies evaluating the same criterion are reported in Sec.~\ref{supp:user_study}.

\subsection{Experimental Setup.}
We evaluate \SIA{} on \flux-Klein{}~\cite{blackforest2025flux2} in concat-mode, with $A$, $A'$, and $B$ as conditions. Both \AnaLoRA{} and \ExLoRA{} use rank $r{=}16$, learning rate $5{\times}10^{-5}$, and up to $50$K steps, with batch sizes of $16$ and $24$ respectively. At inference, we use 28 denoising steps with \ExLoRA{} scale $1.0$. Each \CSV{} direction is estimated from 64 prompt pairs per edit, cached once, and applied with suppression strength $\lambda_i$ (Eq.~\ref{eq:mask_aligned_csv}) selected from $\{0,75,150,225,300\}$. We compare with VisualCloze~\cite{li2025visualcloze}, RelationAdapter~\cite{gong2025relationadapter}, LoRWeB~\cite{manor2026lorweb}, and Edit Transfer~\cite{chen2025edittransfer}.

\subsection{Qualitative Results.}
\begin{figure*}[t]
    \centering
    \includegraphics[width=0.78\linewidth]{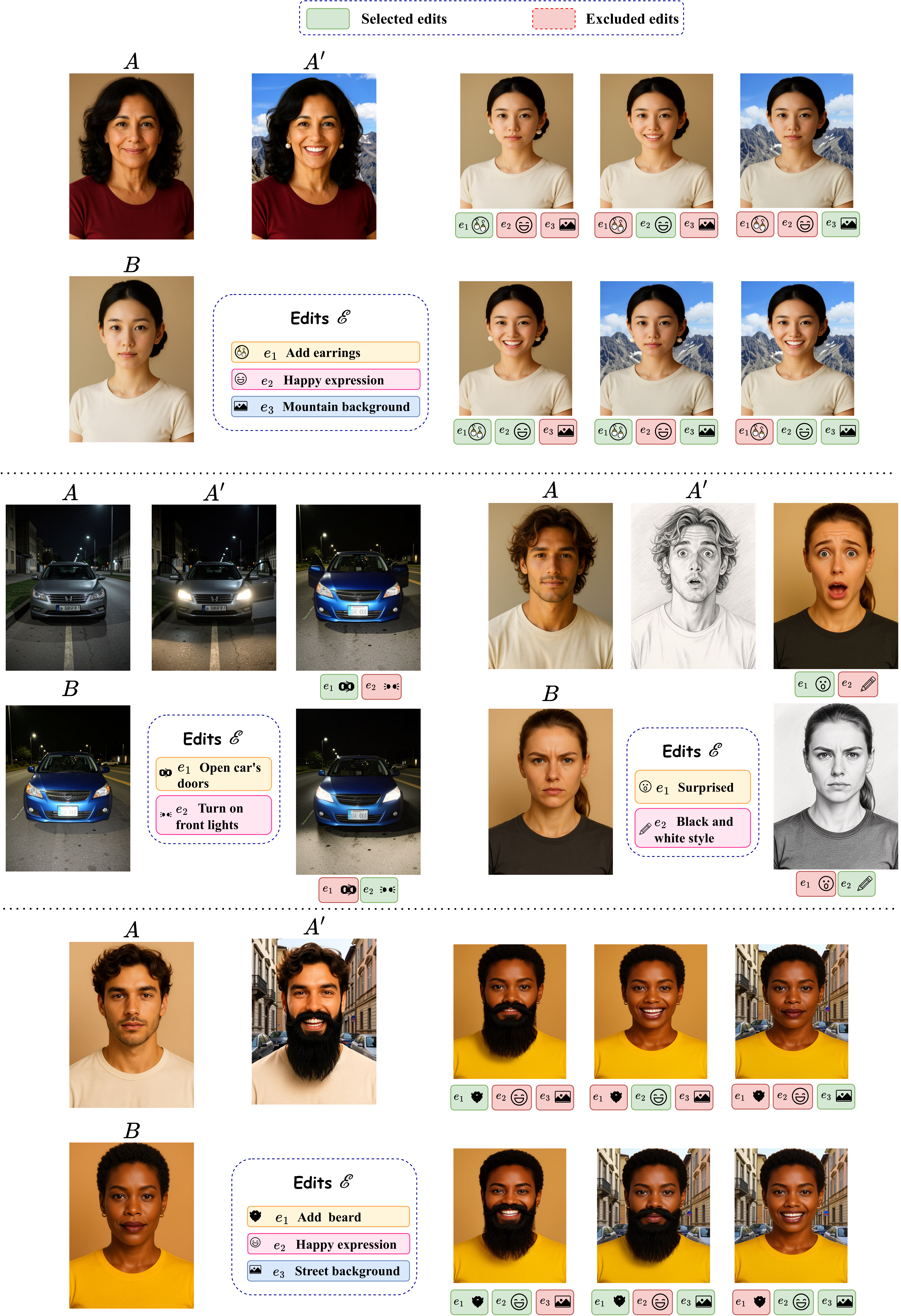}
    \caption{\textbf{Qualitative results.} We provide multiple examples with different numbers of edits that can be switched to active or inactive. Our method is able to successfully control which edit needs to be applied.
    }
    \label{fig:qualitative}
\end{figure*}

Fig.~\ref{fig:qualitative} shows results of \SIA{} across a set of selective image analogy tasks. The examples cover local object additions, facial expression changes, background replacements, and style transfer. Existing visual analogy methods often transfer the full transformation from \(A'\), including edits selected for suppression, or preserve \(B\) by making only weak changes. In contrast, \SIA{} better separates the demonstrated edits: it preserves the retained transformations, suppresses the user-selected edit, and maintains the identity and overall structure of the query image \(B\). These results show that our method can treat a multi-edit demonstration as a controllable set of edit factors rather than a single entangled transformation.

\begin{figure}[!tp]
    \centering
    \includegraphics[width=\linewidth]{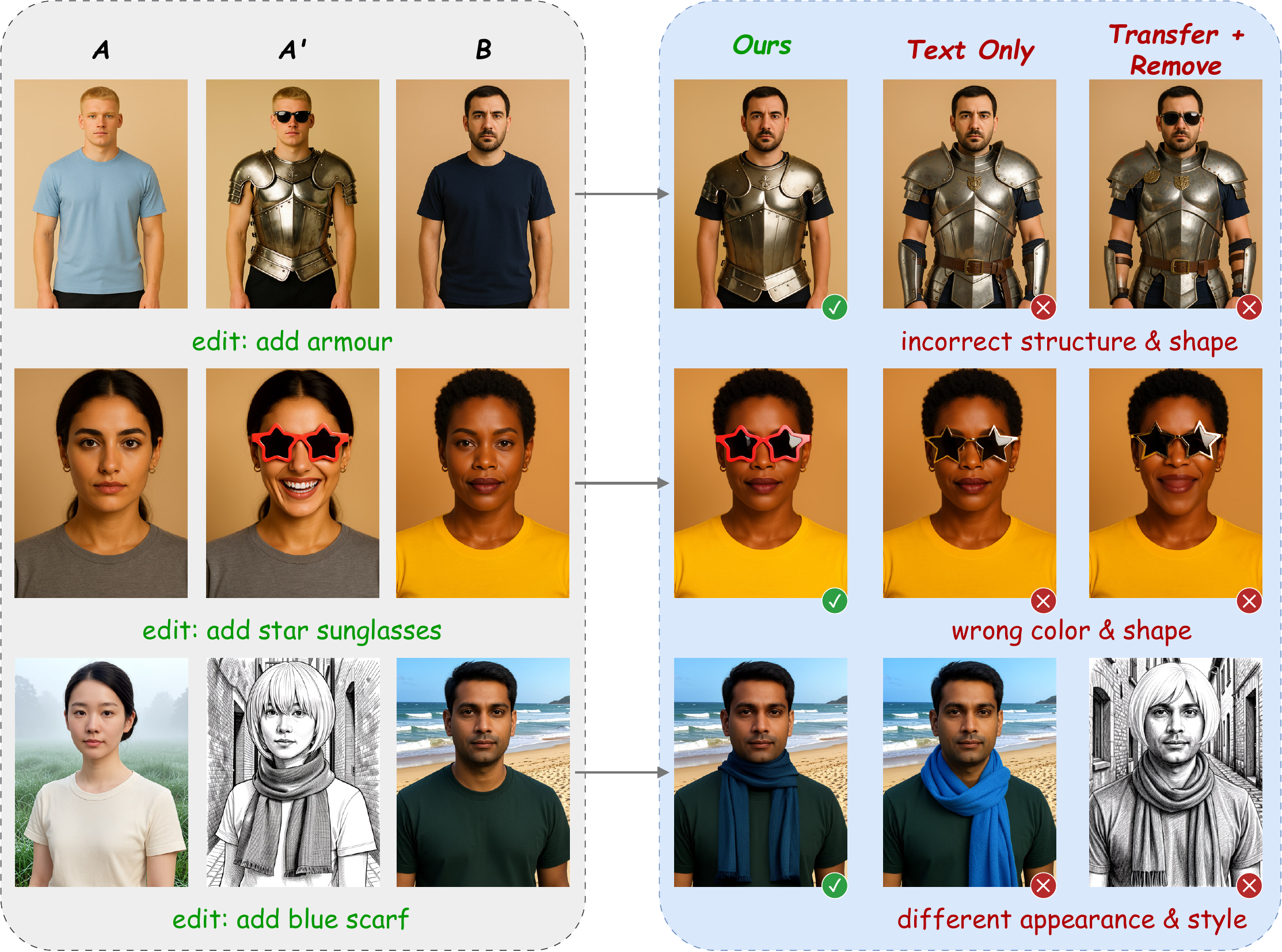}
    \caption{\textbf{Advantage of demonstration-based specification.}
    Text conveys the category of an edit but not its exact form. In each row, the demonstration specifies the exact visual form of the edit, including its structure, shape, and color, which text-based editing fails to reproduce, while SIA matches the demonstrated transformation.}
    \label{fig:sia_vs_textBaseline}
\end{figure}

\begin{figure*}[t]
    \centering
    \includegraphics[width=\linewidth]{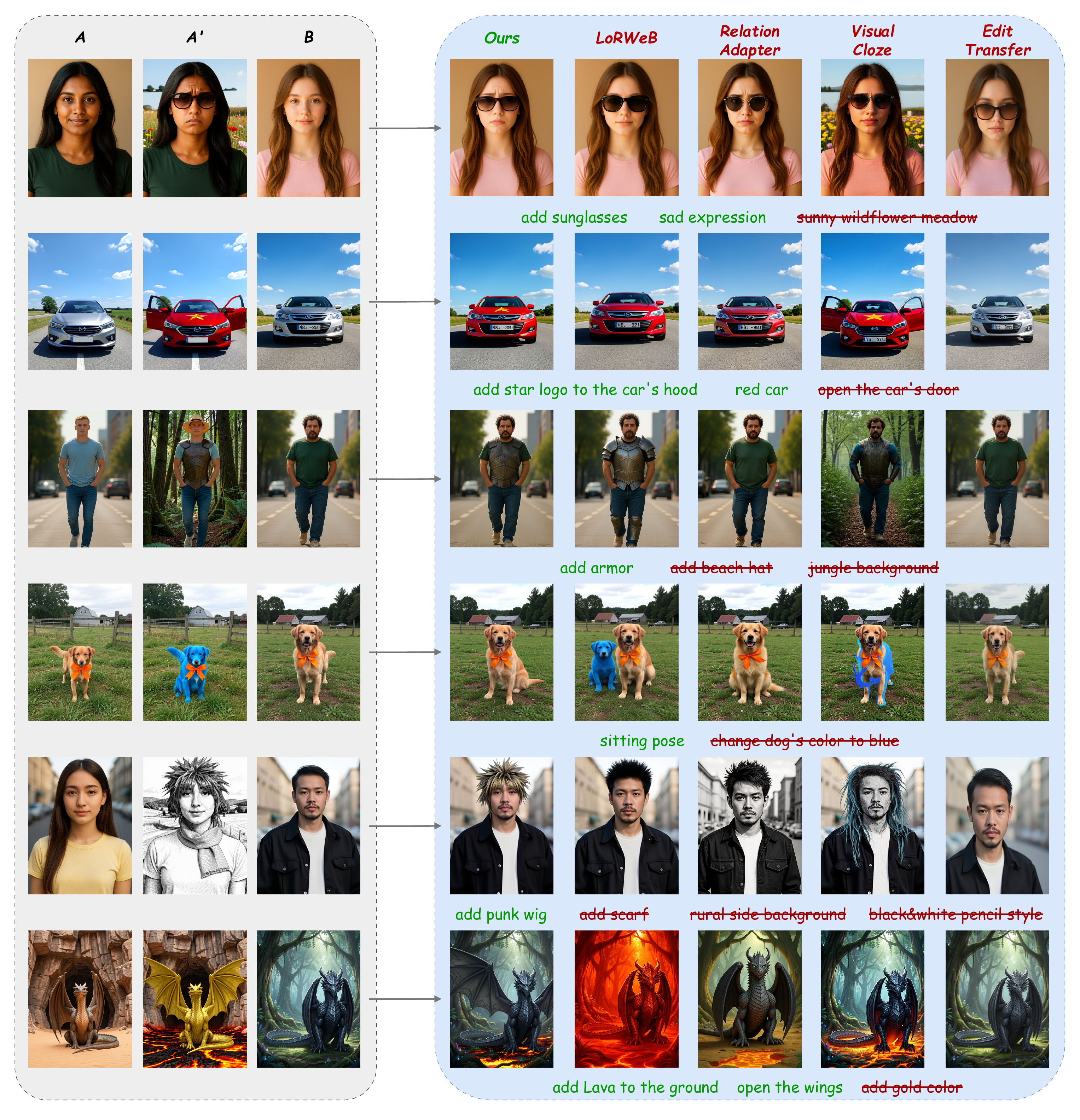}
    \caption{\textbf{Visual comparison.}
   Qualitative comparisons with existing approaches demonstrate that only our method can selectively suppress undesired edits while preserving the input image and applying the intended modifications. In contrast, other methods either apply all edits simultaneously (e.g., Relation Adapter) or substantially alter the input image, such as the dog introduced by LoRWeb.}
    \label{fig:visual_comparison}
\end{figure*}

\noindent\textbf{Comparison to Baselines.}
Since existing visual analogy methods are designed for full edit transfer rather than selective transfer, we adapt them to our setting by providing the same input triplet \((A,A',B)\). For the input text prompt, we provide the retained-edit prompt. This gives each baseline access to the user-desired edit subset, while keeping the visual demonstration fixed. The comparison, therefore, tests whether each method can avoid copying suppressed edits from \(A'\), where the full transformation remains visually present. Fig.~\ref{fig:visual_comparison} shows qualitative comparisons with baselines.

\subsection{Quantitative Evaluation.}
We report both CLIP-based and VLM-based metrics. For CLIP-based evaluation, we report two directional metrics. \emph{CLIP-Dir Image} is the image-image directional metric that measures agreement with the demonstrated image-space transformation. \emph{CLIP-Dir Retained} measures whether the generated output moves in the direction of the retained edit descriptions, making it suitable for selective transfer.

For semantic evaluation, we use a VLM-based protocol, following the recent use of vision-language models for assessing image editing quality and human-aligned preferences~\cite{huang2025imageeditingsurvey}. Qwen3-VL-32B-Instruct~\cite{Qwen3-VL} and Gemma~4~\cite{gemmateam2026gemma4} are used as independent judges; results with Gemma~3~\cite{gemmateam2025gemma3} as a third judge are reported in the supplementary material. Each VLM is given the demonstration pair \((A,A')\), the query image \(B\), the generated output \(\hat{B'}\), and the lists of retained and suppressed edits. We report three VLM metrics. \emph{Retained Accuracy} measures the quality of applying the retained edits. \emph{Preservation} measures whether the output maintains the identity, layout, and unrelated content of \(B\). \emph{Edit Recall} measures the fraction of retained edits detected in the output.

Tab.~\ref{tab:quantitative} compares \SIA{} against recent visual analogy baselines. SIA achieves the highest retained-edit accuracy and edit recall under both VLM judges (and under Gemma~3; see the supplementary material), and also obtains the best CLIP directional scores. Under
Gemma~4, SIA also attains the highest preservation among the analogy baselines.
We note that methods such as Edit Transfer can obtain high preservation because they often make only weak changes or leave the query image close to unchanged. However, preservation alone is not sufficient for selective image analogy: a successful method must also apply the retained edits accurately, which is reflected by Retained Accuracy and Edit Recall.

\noindent\textbf{Comparison to text-based only editing.}
Given the strength of text- and reference-conditioned editors, it is worth examining if a demonstration pair is necessary. Fig.~\ref{fig:sia_vs_textBaseline} demonstrates that SIA better captures the desired edit while avoiding unwanted changes. Using FLUX.2-Klein-9B with a retained-edit text instruction, the \emph{text-only} baseline struggles to suppress unwanted edits, while \emph{Full Transfer + Removal} first copies all demonstrated edits and then attempts to remove the suppressed ones. Additional quantitative and qualitative comparisons with text-only baselines and commercial tools such as Nano Banana are provided in the supplementary material.

\subsection{Ablation Study.}
We ablate the two selective suppression components of \SIA{} in Tab.~\ref{tab:ablation} and the supplementary material (Sec.~\ref{supp:visual_ablation}). In addition to Retained Accuracy and Preservation, we report a VLM-based Suppression score (\emph{Supp.}), measuring whether suppressed edits are absent from the output. \AnaLoRA{} remains active in all variants as the base visual-analogy capability. Without \ExLoRA{} and \CSV{}, the model transfers nearly the full transformation from $A'$. \CSV{} alone weakens the suppressed edit's text signal but lacks a learned suppression pathway, while \ExLoRA{} alone provides token-local suppression but leaves residual semantic evidence in the prompt embeddings. Combining both yields the strongest suppression and overall performance. The slight drop in Retained Accuracy reflects the trade-off between suppression strength and retained-edit fidelity, while visual ablations confirm their complementary roles.

\begin{table}
\caption{Quantitative ablation of the suppression components on the test set. Supp.\ measures the absence of suppressed edits in the output. \AnaLoRA{} is active in all variants.}
\label{tab:ablation}
\begin{tabular}{cccccc}
\toprule
ExLoRA & CSV & Ret. Acc. $\uparrow$ & Supp. $\uparrow$ & Pres. $\uparrow$ & Avg. $\uparrow$ \\
\midrule
No  & No  & \textbf{9.07} & 1.93 & 9.00 & 6.67 \\
No  & Yes & 8.81 & 4.93 & 9.16 & 7.63 \\
Yes & No  & 7.95 & 7.81 & \textbf{9.77} & 8.51 \\
Yes & Yes & 8.01 & \textbf{7.85} & 9.75 & \textbf{8.54} \\
\bottomrule
\end{tabular}
\end{table}
\section{Limitations, Future Work and Conclusion}
\label{sec:conclusion}
We introduced Selective Image Analogy, a novel problem setting that, to our knowledge, is the first to enable selective transfer from a multi-edit visual demonstration. Unlike prior image analogy methods that treat the demonstrated transformation as an indivisible whole, \SIA{} allows a user to choose which edits from a single before/after pair are applied to a new query image, while suppressing the rest---even when the suppressed edits remain visible in the demonstration. To support this setting, we constructed a multi-edit analogy dataset and proposed a mask-aligned suppression mechanism combining \AnaLoRA{} for full-transfer learning, \ExLoRA{} for token-gated suppression during denoising, and \CSV{} for weakening residual edit signals in the text embeddings before denoising. Extensive quantitative evaluations, qualitative comparisons, and a user study confirm that \SIA{} consistently outperforms existing visual analogy baselines on the selective transfer objective, demonstrating that a single demonstration pair can serve as a flexible, user-controllable source of edits rather than a fixed, all-or-nothing transformation.

While \SIA{} currently supports up to a handful of demonstrated edits, it may struggle when $A$ and $A'$ differ substantially. The dataset also covers a limited range of edit types and scenes, leaving out fine-grained operations such as artifact removal and complex layouts. Scaling to a larger number of edits and broader edit diversity remains an interesting direction for future work.
Another promising direction is compositional demonstration, where edits are drawn from multiple different before/after pairs and combined into a single selective transfer, allowing users to mix demonstrated edits across sources with controllable intensity rather than being constrained to a single $(A, A')$.

\bibliographystyle{ACM-Reference-Format}
\bibliography{main}

\clearpage
\clearpage

\appendix

\providecommand{\Rcal}{\mathcal{R}}
\providecommand{\Scal}{\mathcal{S}}
\providecommand{\Ecal}{\mathcal{E}}

\section{Implementation Details}
\label{supp:impl}

\paragraph{Backbone.}
For all experiments, we use the 9B-parameter variant of \flux-Klein{}~\cite{blackforest2025flux2} as the diffusion backbone. The text stream is encoded by the Qwen3 text encoder packaged with \flux-Klein{}. The diffusion pipeline concatenates the hidden states from three Qwen3 layers ($\{9,18,27\}$) along the channel dimension to form the per-token prompt embedding consumed by the MM-DiT blocks. As described in the main paper, \ExLoRA{} is injected into the text-stream projections of the MM-DiT blocks.

\paragraph{Adapters.}
\AnaLoRA{} and \ExLoRA{} are independent LoRA adapters with rank $r{=}16$ and dropout $0$. We first train \AnaLoRA{} with batch size $16$, learning rate $5{\times}10^{-5}$, and up to $50\mathrm{K}$ optimization steps on a mixture of full-transfer tuples from the two-, three-, and four-edit groups. We then merge the resulting checkpoint into the frozen backbone and train \ExLoRA{} on top, using batch size $24$ and the same learning rate. To keep \ExLoRA{} close to the full-transfer behavior at the beginning of training, roughly one third of the optimization steps instead update \AnaLoRA{} jointly with an anchoring weight of $0.1$, with the suppression mask disabled. Both adapters are optimized with AdamW and gradient clipping at $\|g\|_2\le 1$. We use the flow-matching loss described in the main paper, with timesteps sampled uniformly from $[0,1000)$.

\paragraph{Inference.}
At inference time, we run $28$ denoising steps in concat mode. The three conditioning latents $z_A,z_{A'},z_B$ are concatenated along the token dimension, and the model is not given a $2{\times}2$ grid. The Klein backbone does not support classifier-free guidance, so we do not pass a negative prompt. The \ExLoRA{} scale is fixed at $1.0$ for all reported numbers. Setting it to $0$ gives the \AnaLoRA{}-only baseline.

\paragraph{\CSV{} pipeline.}
For every edit phrase, the steering direction is built once and cached. The pipeline follows the Diffusion-Sliders~\cite{ekin2026steering} recipe, but uses the diffusion pipeline's own text encoder so that the direction is defined in exactly the embedding space consumed by the denoiser:
\begin{enumerate}[leftmargin=*,nosep]
    \item For each edit, we prompt the same Qwen3 encoder in causal-LM mode to generate $N{=}64$ contrastive pairs $\{(p_n^+,p_n^-)\}$, where the target edit is the only varying attribute.
    \item We re-encode every sentence with the same encoder to obtain the per-layer hidden states.
    \item We mean-pool the hidden states at the token positions that name the target edit.
    \item We store the unit-normalized difference of means
    \(
        d_i = \frac{\bar{u}_i^+ - \bar{u}_i^-}{\|\bar{u}_i^+ - \bar{u}_i^-\|_2}
    \)
    on disk together with the generated pairs. The cache is keyed by $\mathrm{sha1}(\text{edit phrase})$, so the same vector is reused across all queries that mention the same edit.
    \item At inference, every token of a suppressed edit phrase is updated with Eq.~7 of the main paper: its component along $d_i$ is removed and it is moved by $-\lambda_i d_i$. Tokens of retained edits and of the analogy instruction are left unchanged.
\end{enumerate}

\paragraph{Contrastive pair construction.}
Following the dataset prompt of Diffusion Sliders~\cite{ekin2026steering}, we ask the encoder for $N{=}64$ pairs in JSON form. Each pair has a positive and a negative sentence and the identifier words that differ between them. The two sentences share the same subject, setting, and composition, and differ only in the words that express the edit, for example ``\emph{A person with a beard sitting in a park}'' versus ``\emph{A clean-shaven person sitting in a park}''. Subjects, settings, and wording vary across pairs, so the direction is not tied to one context. We sample with temperature $1.0$ and nucleus sampling ($p{=}0.95$). Since the embedding is pooled at the identifier tokens, we discard any pair whose identifier does not appear verbatim in its sentence. In the rare case that fewer than $\max(4, N/2)$ pairs remain, we fill the set with simple template pairs such as ``\emph{A photo of a subject showing $\langle$edit$\rangle$}'' versus ``\emph{A photo of a subject, unchanged}''.

\paragraph{Hardware and training time.}
All training runs were performed on a single \texttt{RTX PRO 6000 WS} GPU. Inference for one $(A,A',B)$ tuple at $28$ steps and $512{\times}512$ resolution takes under one minute. We will release the training configuration files and evaluation scripts to support reproducibility.

\section{User Study}
\label{supp:user_study}

\begin{table}[t]
\centering
\caption{
\textbf{User study results}. Participants rank the top three outputs according to selective analogy quality. Our method receives the highest average ranking score.
}
\label{tab:user_study}
\begin{tabular}{lr}
\toprule
Method & Ave. Ranking Score \(\uparrow\) \\
\midrule
RelationAdapter~\cite{gong2025relationadapter} & 1.22 \\
VisualCloze~\cite{li2025visualcloze} & 0.81 \\
Edit Transfer~\cite{chen2025edittransfer} & 0.12 \\
LoRWeB~\cite{manor2026lorweb} & 1.38 \\
\midrule
\emph{\textbf{Ours}} & \textbf{2.47} \\
\bottomrule
\end{tabular}
\end{table}

\begin{table}[t]
\centering
\caption{
\textbf{Text-baseline user study.} Participants rate each output for Fidelity on a $0$--$10$ scale and choose the single best overall result per sample. \SIA achieves the highest Fidelity and is most often chosen as best overall.
}
\label{tab:user_study_text}
\begin{tabular}{lcc}
\toprule
Method & Fidelity $\uparrow$ & Overall (\%) $\uparrow$ \\
\midrule
Text Only               & 7.00 & 30.1 \\
Full Transfer + Removal & 5.78 &  4.4 \\
\midrule
\emph{\textbf{SIA (Ours)}} & \textbf{8.86} & \textbf{65.4} \\
\bottomrule
\end{tabular}
\end{table}

Standard image-level metrics, such as LPIPS and CLIP similarity, do not fully capture the main requirement of selective image analogy, where the output must apply the retained edits, avoid the suppressed edit, and still preserve the query image \(B\). We therefore conduct a user study focused on this task-specific criterion. In each trial, participants are shown the demonstration pair \((A,A')\), the query image \(B\), the retained edits, the edit selected for suppression, and outputs from \SIA and the baseline methods in random order. Participants are asked to rank the top three outputs according to how well they satisfy the selective analogy objective, considering three factors jointly: applying the retained edits, avoiding the suppressed edit, and preserving the identity and structure of \(B\). We collect responses from \(37\) participants over \(15\) examples and compare \SIA against the baselines. For scoring, the first-ranked result receives \(3\) points, the second receives \(2\) points, the third receives \(1\) point, and unranked results receive \(0\) points. The results in Tab.~\ref{tab:user_study} show that our method receives the highest average ranking score, confirming that \SIA better matches the intended selective edit behavior.

We additionally conduct a second user study that directly compares \SIA against the \emph{Text Only} and \emph{Full Transfer + Removal} baselines, the strongest text-conditioned alternatives to demonstration-based transfer. In each trial, participants see the demonstration pair \((A,A')\), the query image \(B\), the retained-edit and suppressed-edit lists, and the outputs of the three methods in random order. Participants rate the \emph{Fidelity} on a \(0\)--\(10\) scale. \emph{Fidelity} measures whether the retained edits match the demonstrated form in \(A'\) with the
same style, shape, and color, and not merely the same category of edit, where a low score means the retained edits are missing or bear no resemblance to \(A'\), a
mid-range score means the correct type of edit but a different instance (for example, sunglasses of a different style), and the maximum means every retained edit is present and matches the exact appearance in \(A'\). After rating all three outputs, participants additionally select the single best overall result for that sample. We collect responses from \(20\) participants over \(8\)
examples. As shown in Tab.~\ref{tab:user_study_text}, \SIA attains the highest Fidelity score and is chosen as the best overall result in \(65.4\%\) of trials, compared with \(30.1\%\) for Text Only and \(4.4\%\) for Full
Transfer + Removal. While text-based editing often produces the correct semantic edit, it fails to reproduce the demonstrated visual form, which is precisely what
the Fidelity metric penalises.


\section{Comparison to text-based editing and commercial models}
\label{sec:Comparisontext-based}

A natural question is whether we even need the demonstration pair at all, given how strong text- and reference-conditioned editors have become. To check this, we compare SIA against three additional
baselines on the same test set. \emph{(i) Text Only} simply runs FLUX.2-Klein-9B with the retained-edit instruction as text, with no demonstration pair involved. \emph{(ii) Full Transfer + Removal} first copies over every demonstrated edit, then tries to undo the suppressed ones with a text instruction. \emph{(iii) Nano Banana} is a commercial black-box editor that takes the full triplet $(A, A', B)$ along with the retained-edit instruction; qualitative comparisons against it are shown in Sec.~\ref{supp:sia_vs_gemini}. \\
Results are reported in Tab.~\ref{tab:text_baselines_qwen}.
All three baselines score high on Recall, meaning they generally produce the right \emph{kind} of edit. But Recall only checks whether an edit happened. It does not check whether the edit looks like the one demonstrated in $A'$, matching its style, shape, size, placement, and scale. That is what Retained Accuracy measures, and this is where \SIA pulls ahead of every baseline (Fig.~\ref{fig:sia_vs_textBaseline}). \emph{Full Transfer + Removal} does even worse. Its removal step often changes parts of $B$ that should have stayed the same, which lowers its Preservation score.
These metrics can still miss subtler visual mismatches, so we run a second check: a head-to-head comparison judged by Qwen3-VL-32B-Instruct (Tab.~\ref{tab:pairwise}). For each pair, the judge sees $(A, A', B)$, the retained-edit description, and one output from \SIA{} next to one from either \emph{Text Only} or \emph{Full Transfer + Removal}. The judge then picks the image that better applies the retained edit while suppressing the rest. Across both comparisons, \SIA{} is chosen as better or tied roughly 70\% of the time.

Tab.~\ref{tab:text_baselines_gemma4} repeats the comparison with Gemma~4 as an independent judge. The trend holds: SIA leads on Retained Accuracy by a full point over both text-based baselines, and also attains the highest Recall.

\begin{table}[t]
\caption{Comparison with text-based editing baselines and a commercial
multi-reference editor, judged by Qwen3-VL-32B-Instruct. Text-based baselines
achieve high Recall but lower Retained Accuracy.}
\label{tab:text_baselines_qwen}
\small
\begin{tabular}{lccc}
\toprule
Method & Ret. Acc.$\uparrow$ & Recall$\uparrow$ & Pres.$\uparrow$ \\
\midrule
Text Only               & 7.43 & 0.917 & 9.68 \\
Full Transfer + Removal & 7.68 & \textbf{0.921} & 8.99 \\
Nano Banana             & 7.36 & 0.885 & 9.55 \\
\midrule
\textbf{SIA (Ours)}     & \textbf{8.01} & 0.892 & \textbf{9.75} \\
\bottomrule
\end{tabular}
\end{table}
 
\begin{table}[t]
\caption{Comparison with text-based editing baselines, judged by Gemma~4.
SIA attains the highest Retained Accuracy and Recall, while Text Only obtains
the highest Preservation.}
\label{tab:text_baselines_gemma4}
\small
\begin{tabular}{lccc}
\toprule
Method & Ret. Acc.$\uparrow$ & Recall$\uparrow$ & Pres.$\uparrow$ \\
\midrule
Text Only               & 8.31 & 0.941 & \textbf{9.95} \\
Full Transfer + Removal & 8.36 & 0.949 & 9.71 \\
\midrule
\textbf{SIA (Ours)}     & \textbf{9.31} & \textbf{0.965} & 9.89 \\
\bottomrule
\end{tabular}
\end{table}

\begin{table}
\caption{Side-by-side VLM preference. The judge sees $(A,A')$, $B$, the retained/suppressed edit lists, and two outputs in random order.}
\label{tab:pairwise}
\resizebox{\linewidth}{!}{%
\begin{tabular}{lccc}
\toprule
Baseline & Ours Win (\%) $\uparrow$ & Baseline Win (\%) $\downarrow$ & Tie \% \\
\midrule
Text Only               & \textbf{56.9} & 17.6 & 25.5 \\
Full Transfer + Removal & \textbf{48.5} & 30.1 & 21.4 \\
\bottomrule
\end{tabular}%
}
\end{table}

\section{SIA vs Nano Banana qualitative comparison}
\label{supp:sia_vs_gemini}

\begin{figure}[t]
    \centering
    \includegraphics[width=\linewidth]{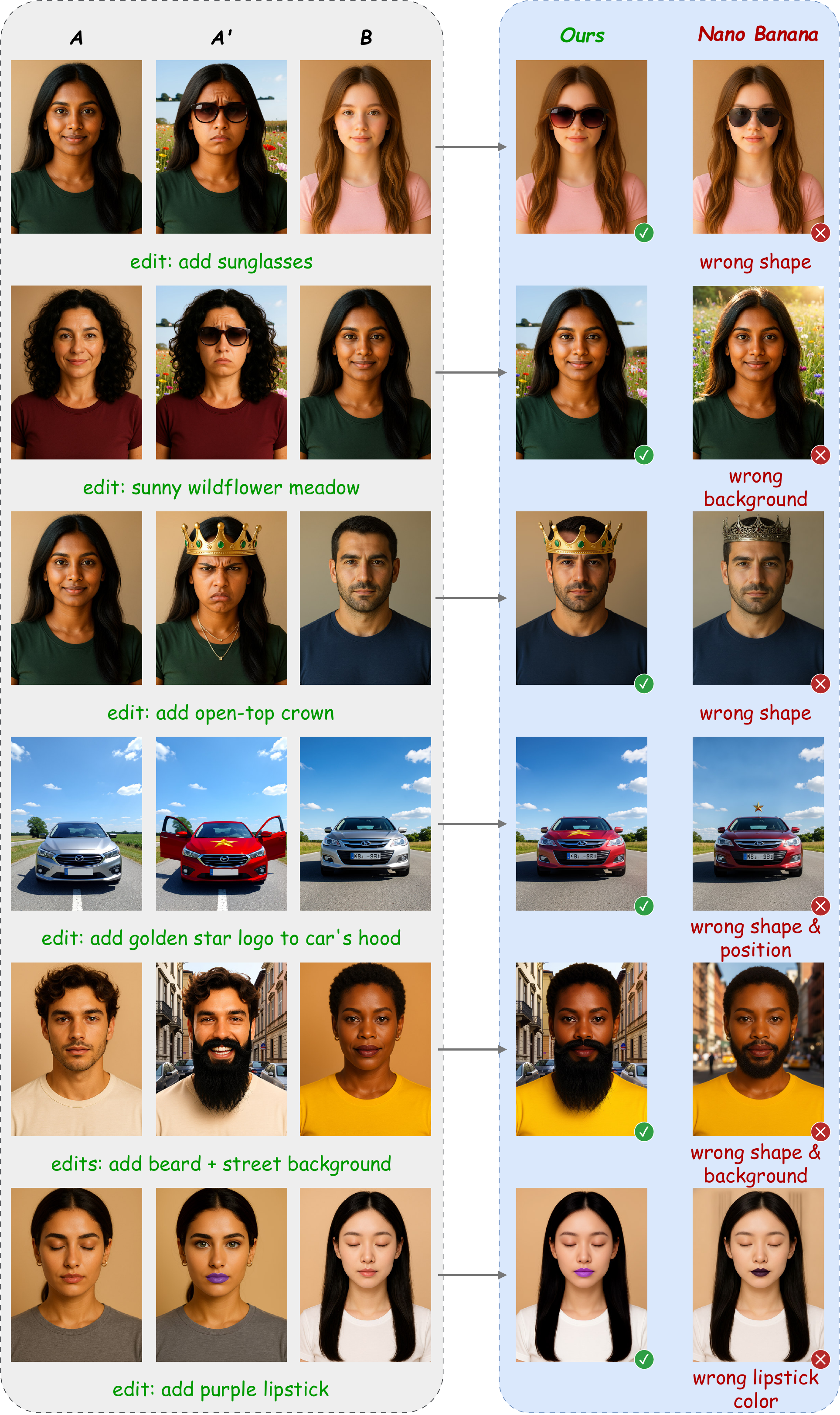}
\caption{\textbf{Visual comparison with Nano Banana.}
   Qualitative comparisons against Nano Banana, a commercial multi-reference editor, show that only \SIA{} reproduces the demonstrated edit's exact visual form. Nano Banana typically identifies the correct \emph{category} of edit, but fails to match the specific appearance shown in $A'$: it renders sunglasses and the crown with the wrong shape, places the star logo in the wrong position, and produces an incorrect lipstick color. In contrast, \SIA{} consistently matches both the type and the precise form of each demonstrated edit while preserving the identity and structure of $B$.}
\label{fig:sia_vs_gemini}
\end{figure}

Beyond the analogy-based baselines evaluated in the main paper, a natural question is how \SIA{} compares against strong commercial editors that accept multiple reference images directly. We therefore compare against Nano Banana, a
black-box multi-reference editor, giving it the same demonstration pair $(A,A')$, the query image $B$, and the retained-edit instruction. Fig.~\ref{fig:sia_vs_gemini} shows the results across a range of edits, including adding sunglasses, a crown, a logo, a beard, and lipstick. In every case, Nano Banana recognizes the correct kind of edit but does not reproduce the specific form demonstrated in $A'$: the sunglasses and crown take on the wrong shape, the star logo is placed incorrectly, and the lipstick appears in the wrong color. \SIA{}, by contrast, matches both the type and the appearance of each demonstrated edit while leaving the identity and structure of $B$ intact. This highlights a key advantage of demonstration-based transfer: text or reference conditioning alone conveys
\emph{what} to edit, but not the precise visual form the edit should take.

\section{Multi-Edit Analogy Dataset}
\label{supp:dataset}

\begin{figure*}[t]
    \centering
    \includegraphics[width=\textwidth]{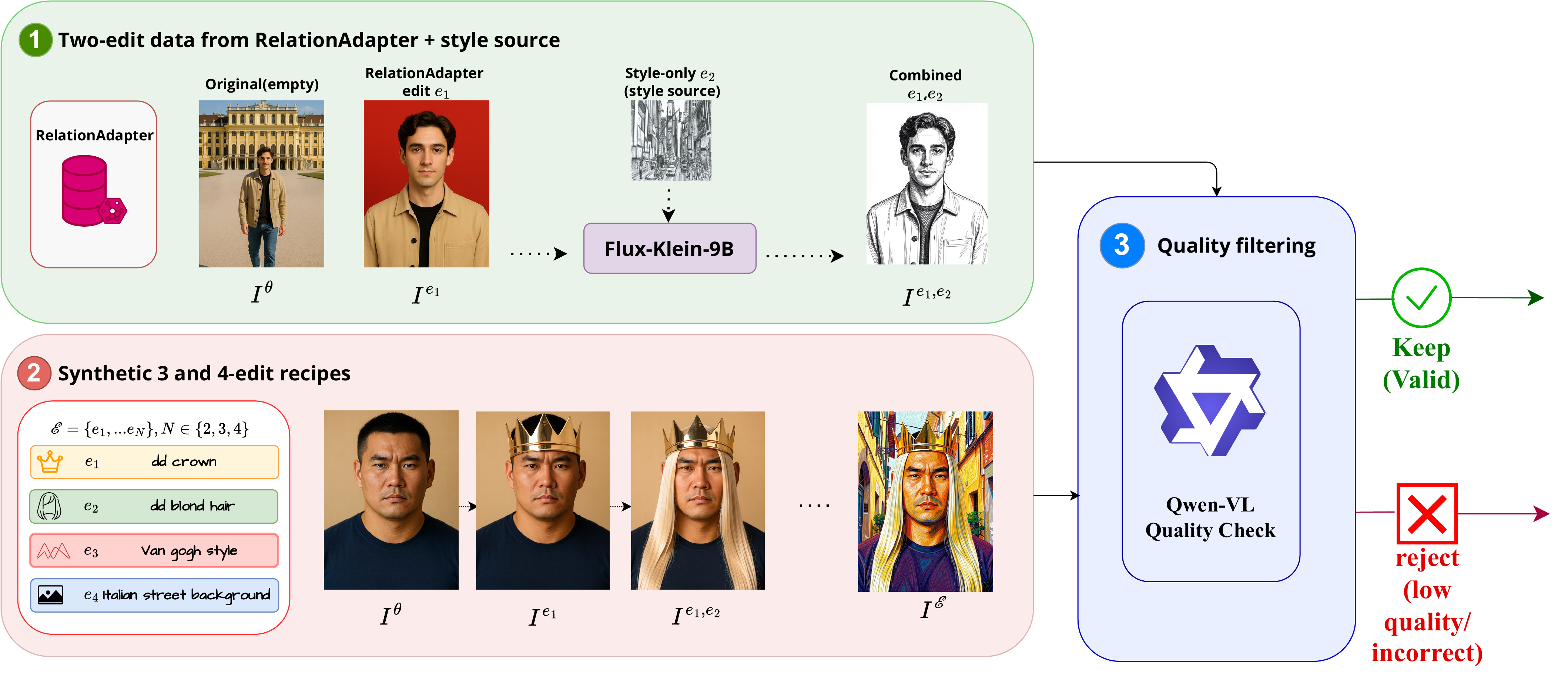}
    \caption{
        \textbf{Overview of our multi-edit dataset generation pipeline.} For higher-order edit settings, we start from a base image and apply edits sequentially to obtain intermediate composites and the final full-edit image. For two-edit settings, we augment RelationAdapter pairs with an additional style transformation synthesized by \flux-Klein. All generated samples are filtered with Qwen2.5-VL to verify that the intended edits are correctly expressed.
    }
    \label{fig:dataset_overview}
\end{figure*}

In multi-edit image analogy, the model must transfer a demonstrated transformation from one subject to another, while also separating that transformation into its individual edit components. This requires supervision beyond standard image-editing datasets. For each subject, we need the original image, the fully edited image, and partial-edit images that correspond to different \emph{subsets} of the demonstrated edits. Since no existing dataset provides this structure, we build a new \textbf{multi-edit analogy dataset} for our task. Fig.~\ref{fig:dataset_overview} shows the generation pipeline, and Fig.~\ref{fig:data_format} shows the storage layout for each sample.

\subsection{Edit recipes and storage format}
\label{supp:dataset:format}

Our dataset is organized around \emph{edit recipes}. An edit recipe defines the set of edits applied to all subjects in a group, such as adding an object, changing the background, or changing the image style. We represent each recipe as a set of edits $\Ecal=\{e_1,\ldots,e_N\}$, where $N\in\{2,3,4\}$. For each subject in a recipe, we store the original image $I^{\emptyset}$, the full-edit image $I^{\Ecal}$, and the intermediate partial images $I^{\Rcal}$ for the retained edit subsets $\Rcal\subseteq\Ecal$ that are reachable through the sequential synthesis process. As in the main paper, $\Rcal$ denotes the retained edits and $\Scal=\Ecal\setminus\Rcal$ denotes the suppressed edits. Each edit also has a short natural-language description, which is used both for prompting and for constructing the contrastive pairs used by \CSV{} (Sec.~\ref{supp:impl}).

Training tuples are formed by pairing two subjects $a$ and $b$ from the same recipe:
\[
    A = I_a^{\emptyset}, \qquad
    A' = I_a^{\Ecal}, \qquad
    B = I_b^{\emptyset}.
\]
For the query subject $b$, the full target $B'_{\Ecal}=I_b^{\Ecal}$ contains all demonstrated edits, while the selective target $B'_{\Rcal}=I_b^{\Rcal}$ contains only the retained edits. Together, these targets provide the supervision needed for selective analogy. The full target $B'_{\Ecal}$ teaches the model to transfer the complete demonstrated transformation, while the selective target $B'_{\Rcal}$ teaches the model how the same demonstration should be interpreted when the mask excludes the edits in $\Scal$.

\begin{figure*}[t]
    \centering
    \includegraphics[width=0.95\textwidth]{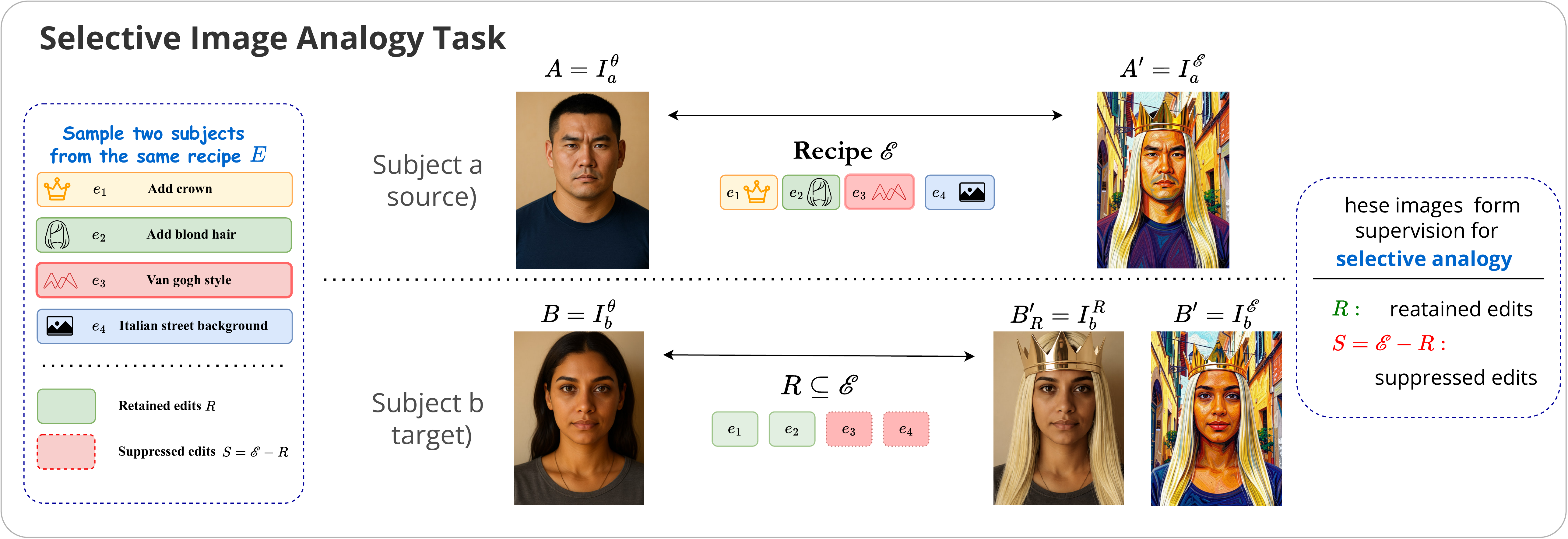}
    \caption{\textbf{Per-sample storage layout.} Each leaf directory stores the source image, one reference image for each edit phrase, the sequence of composite images corresponding to retained subsets $\Rcal\subseteq\Ecal$, and a \texttt{prompt.json} file that maps every composite to a short natural-language description.}
    \label{fig:data_format}
\end{figure*}

Concretely, every leaf directory contains the source image $I^{\emptyset}$, the reference images that visually anchor each individual edit (for example, the style reference, object reference, or target expression), the sequence of composite images corresponding to partial-edit subsets, and a \texttt{prompt.json} file that maps each composite to a short natural-language instruction. Fig.~\ref{fig:data_format} summarizes this layout.

\subsection{Two-edit data from RelationAdapter and style sources}
\label{supp:dataset:two}

The first part of our dataset contains two-edit analogies built from a subset of RelationAdapter~\cite{gong2025relationadapter}. RelationAdapter provides diverse single-edit transformations, such as changing character attributes, modifying pose, adding accessories, or altering local appearance. We treat each selected RelationAdapter pair as the first edit, and introduce a second edit by applying a style transformation defined by a small set of predefined style reference images. We use \flux-Klein{} in style-transfer mode to synthesize both the style-only composite and the final two-edit composite. Each example therefore contains four images of the same subject: the original image, the RelationAdapter-edited image, the style-only image, and the final image that combines both edits. Since RelationAdapter provides most of the visual diversity, this subset is the main source of edit variation in our dataset.

\subsection{Synthetic three-edit and four-edit data}
\label{supp:dataset:multi}

The second part of our dataset contains more complex edit recipes with three or four edits. These examples are designed to expose the model to cases where multiple edits must be suppressed at the same time while the remaining edits are preserved. For these settings, we first generate source images with \flux-Klein{} and then use the same model to synthesize the edited variants, including object additions, style changes, and background changes.

Starting from the original image $I^{\emptyset}$, edits are applied sequentially. We first generate the single-edit image, then the two-edit image, and finally the full multi-edit image. For a three-edit recipe, this gives the progression
\[
    I^{\emptyset} \rightarrow I^{\{e_1\}} \rightarrow I^{\{e_1,e_2\}} \rightarrow I^{\Ecal},
\]
and for a four-edit recipe,
\[
    I^{\emptyset} \rightarrow I^{\{e_1\}} \rightarrow I^{\{e_1,e_2\}} \rightarrow I^{\{e_1,e_2,e_3\}} \rightarrow I^{\Ecal}.
\]
The three-edit group contains recipes such as one object addition with a style change and a background change, two object additions with a style change, or two object additions with a background change. The four-edit group contains a single recipe that combines two object additions, one background change, and one style change. The higher-order groups are less diverse than the two-edit subset, but they serve a different purpose: they teach the model to handle settings where several edits must be suppressed together. In short, the two-edit data provides most of the edit diversity, while the three- and four-edit groups encourage the same suppression behavior in more complex multi-edit cases.

\subsection{Vision-language filtering}
\label{supp:dataset:filter}

The sequential synthesis pipeline can introduce artifacts. An intended edit may fail to appear in a composite, or an edit from an earlier step may be partially undone by a later step. To remove these failure cases, we filter every leaf directory with a vision-language model. For each composite $I^{\Rcal}$, we show the model the source image and the composite, then ask it to check two conditions: (i) every edit in $\Rcal$ is present, and (ii) every edit in $\Ecal\setminus\Rcal$ is absent. Samples that fail either condition are discarded. We use Qwen2.5-VL-32B-Instruct for this filtering step, with separate text-only and image-conditioned passes when needed. The text-only pass is used to paraphrase the high-level transformation label, while the image-conditioned pass is used to verify style or appearance changes. The resulting dataset therefore provides not only the full transformation between $A$ and $A'$, but also the intermediate partial composites needed for selective supervision. It is also filtered so that the desired edits are present and the excluded edits are absent.

\subsection{Dataset Statistics and Composition Analysis}
\label{supp:dataset:stats}

This section describes how the dataset is composed and how it is split into training and test sets. In total, it contains \textbf{1{,}845} samples, organized into \textbf{5} categories and \textbf{238} distinct edit-recipe \emph{settings}. Each category is defined by two properties: the number of edits in the recipe, $N \in \{2,3,4\}$, and whether the subject is a stylized \emph{creature} (the default, majority case) or a \emph{non-creature} object such as a vehicle or prop. Across these categories, the dataset spans five fundamental edit types: object addition, background change, style transfer, pose change, and facial expression change. Tab.~\ref{tab:dataset_stats} reports the full composition.

\begin{figure*}[p]
    \centering
    \includegraphics{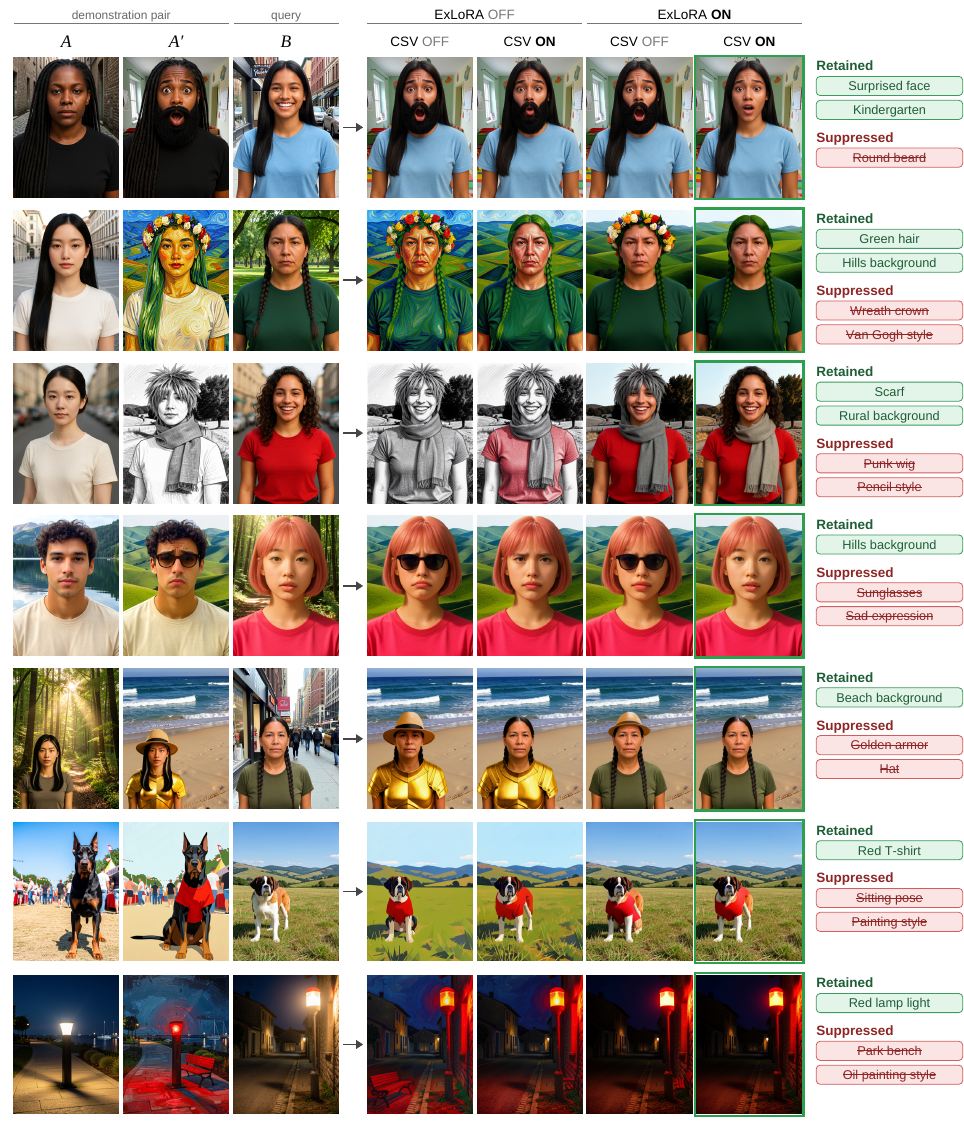}
\caption{\textbf{Visual ablation.} Each row shows one demonstration pair $(A,A')$ and query $B$, with the four combinations of \ExLoRA{} and \CSV{} as columns (see the key at the top); the full method is boxed in green. The edits to retain (green) and to suppress (struck-through red) are listed on the right. Note that \AnaLoRA{} is always active as it is needed to capture all the provided edits. All outputs are \flux-Klein{} results with an \ExLoRA{} trained for only $250$ steps, on training-subject cases, with \CSV{} as an additive push; the examples are selected (see text).}
\label{fig:visual_ablation}
\end{figure*}

\begin{table*}[t]
    \centering
    \caption{\textbf{Dataset composition and train/test split.} Each category is defined by its edit count $N$ and by whether the subject is a creature or a non-creature object. Settings is the number of distinct edit recipes in that category. Train~\% and Test~\% are shares within the category; \% of Data is the category's share of all 1{,}845 samples. Subtotal rows aggregate categories with the same $N$.}
    \label{tab:dataset_stats}
    \small
    \begin{tabular}{lccrrrrrrr}
        \toprule
        Category & $N$ & Variant & Settings & Train & Test & Total & Train \% & Test \% & \% of Data \\
        \midrule
        2edits                  & 2 & Creature     & 210 & 1{,}459 & 19  & 1{,}478 & 98.71\% & 1.29\%  & 80.11\% \\
        2edits non creature     & 2 & Non-creature & 3   & 21      & 36  & 57      & 36.84\% & 63.16\% & 3.09\%  \\
        \cmidrule(l){1-10}
        \emph{Subtotal}         & 2 & --           & 213 & 1{,}480 & 55  & 1{,}535 & 96.42\% & 3.58\%  & 83.20\% \\
        \midrule
        3edits                  & 3 & Creature     & 16  & 62      & 88  & 150     & 41.33\% & 58.67\% & 8.13\%  \\
        3edits non creature     & 3 & Non-creature & 4   & 46      & 72  & 118     & 38.98\% & 61.02\% & 6.40\%  \\
        \cmidrule(l){1-10}
        \emph{Subtotal}         & 3 & --           & 20  & 108     & 160 & 268     & 40.30\% & 59.70\% & 14.53\% \\
        \midrule
        4edits                  & 4 & Creature     & 5   & 18      & 24  & 42      & 42.86\% & 57.14\% & 2.28\%  \\
        \midrule
        \textbf{Total}          & -- & --          & \textbf{238} & \textbf{1{,}606} & \textbf{239} & \textbf{1{,}845} & \textbf{87.05\%} & \textbf{12.95\%} & \textbf{100.00\%} \\
        \bottomrule
    \end{tabular}
\end{table*}

\paragraph{Train/test partitioning.}
Overall, \textbf{1{,}606} samples (87.05\%) are used for training and \textbf{239} (12.95\%) for testing, a ratio of roughly $6.72{:}1$. This split is not applied uniformly across categories. Instead, the ratio in each category reflects the role that category plays in the dataset.

The 2edits creature category is by far the largest and most varied part of the dataset, with 1{,}478 samples across 210 settings. It supplies most of the visual and recipe diversity, so we use almost all of it for training (98.71\%) and hold out only 19 samples for testing. Every other category is majority-test: 63.16\% for 2edits non-creature, 58.67\% for 3edits, 61.02\% for 3edits non-creature, and 57.14\% for 4edits. The reasoning is simple. The large two-edit creature pool provides the training signal, while the smaller and more narrowly built categories are reserved mainly for evaluation. They let us test whether the model's selective-suppression behavior carries over to more complex recipes and to a different type of subject, rather than adding much training signal of their own.

The settings counts follow the same pattern. The 2edits category alone accounts for 210 of the 238 settings (88.3\%), yet contributes only 19 test samples. The remaining four categories together contribute just 28 settings (11.7\%), but send most of their samples to the test split.

\paragraph{Distribution across edit complexity.}
It is also useful to group samples purely by edit count, ignoring the creature and non-creature distinction. The subtotal rows of Tab.~\ref{tab:dataset_stats} give this view. Two-edit recipes dominate at 83.20\% of all samples. This matches our design: the RelationAdapter-derived two-edit subset (Sec.~\ref{supp:dataset:two}) supplies most of the visual and semantic diversity, while the three- and four-edit subsets (Sec.~\ref{supp:dataset:multi}) are deliberately smaller and built specifically to exercise multi-edit suppression, where two or more edits must be excluded at once.

As a result, the three-edit tier sends a much larger share of its samples to the test split (160/268, 59.7\%) than the two-edit tier does (55/1{,}535, 3.6\%), with the four-edit tier close behind at 24/42 (57.1\%). The large two-edit pool teaches the model the underlying structure of edit interactions rather than letting it memorize specific edit combinations. The three- and four-edit samples then extend this to cases where several unwanted edits must be suppressed together. Together, the two give a balanced range of editing complexity, so the model generalizes across different numbers of edits while still suppressing edits reliably.

\paragraph{Approach for training with dataset.}
We build training instances by pairing samples. For each setting in the training split, we form every possible pair of samples within that setting, assigning one sample to the analogy source $(A, A')$ and the other to the analogy target $(B, B')$, together with the edit descriptions for that setting.

Each pair then yields several training instances, one for each retained subset available in that category (Sec.~\ref{supp:dataset:format}). A pair from a 2-edit setting yields 3 instances
($I^{\{e_1\}}$, $I^{\{e_2\}}$, $I^{\mathcal{E}}$), while a pair from a
2-edit non-creature setting yields 2
($I^{\{e_1\}}$, $I^{\mathcal{E}}$). A pair from a 3-edit or 3-edit
non-creature setting yields 3
($I^{\{e_1\}}$, $I^{\{e_1,e_2\}}$, $I^{\mathcal{E}}$), and a pair from a
4-edit setting yields 4
($I^{\{e_1\}}$, $I^{\{e_1,e_2\}}$, $I^{\{e_1,e_2,e_3\}}$, $I^{\mathcal{E}}$). Applied across all training settings, this produces 23{,}992 effective training instances, which we find sufficient for our selective image analogy setting. Tab.~\ref{tab:pairing_calc} summarizes this calculation.

\begin{table}[t]
    \centering
    \caption{\textbf{Same-setting pairing calculation for training-instance construction.} For each category, we sum $\binom{M}{2}$ over all training settings, where $M$ is the number of training samples in a given setting, then multiply by the per-category edit multiplier to obtain the effective training instances. Per-setting sample counts $M$: 2-edit non-creature $= 7,8,6$; 3-edit, avg.\ $M \approx 3.9$; 3-edit non-creature $= 10,12,13,11$; 4-edit $= 3,2,4,4,5$.}
    \label{tab:pairing_calc}
    \small
    \begin{tabular}{lrrrr}
        \toprule
        Category & Settings & Num.\ pairs & Multiplier & Instances \\
        \midrule
        2-edit               & 210 & 7{,}572 & 3 & 22{,}716 \\
        2-edit non-creature  & 3   & 64    & 2 & 128    \\
        3-edit               & 16  & 104   & 3 & 312    \\
        3-edit non-creature  & 4   & 244   & 3 & 732    \\
        4-edit               & 5   & 26    & 4 & 104    \\
        \midrule
        Total & 238 & -- & -- & 23{,}992 \\
        \bottomrule
    \end{tabular}
\end{table}

\section{Visual Ablation}
\label{supp:visual_ablation}
We qualitatively ablate the two selective suppression components of \SIA{} in Fig.~\ref{fig:visual_ablation}.

Fig.~\ref{fig:visual_ablation} shows seven examples, each rendered with the four combinations of \ExLoRA{} and \CSV{} (\AnaLoRA{} is always active). To make the effect of each component visible, the examples come from a separate, shorter training run rather than from the model of the main experiments: \AnaLoRA{} is trained for 2K steps at the native $384{\times}512$ resolution of our data, and \ExLoRA{} is trained from it for only 250 steps. After 500 steps, \ExLoRA{} already suppresses almost perfectly on these cases, and the four variants become hard to tell apart. The cases are training subjects from a development set of 65 cases, rendered with one seed, and \CSV{} is applied as an additive push with $k{=}1$ rather than the projection of Sec.~\ref{sec:method}. We selected rows in which the full model is correct and at least one other variant visibly fails; in some rows another variant is also nearly correct. Over all 494 items of the development set, the mean judge score (Qwen3-VL-32B-Instruct, average of Retained Accuracy, Suppression and Preservation on a 0--10 scale) is $7.90$ for \AnaLoRA{} only, $8.47$ with \CSV{}, $8.77$ with \ExLoRA{} and $9.34$ with both. Since both adapters have seen these subjects during training, only the differences between the variants are meaningful. The gain from \CSV{} disappears as \ExLoRA{} trains longer: after 500 steps, \ExLoRA{} alone reaches $9.68$, and adding \CSV{} changes the average by only $+0.005$ (sign test $p{=}0.42$). This matches Tab.~\ref{tab:ablation}, where adding \CSV{} to the fully trained \ExLoRA{} changes the average only from $8.51$ to $8.54$, while \CSV{} without \ExLoRA{} improves it by $0.96$.

\section{Evaluation Metric Details}
\label{supp:metrics}
Selective image analogy places an unusual demand on evaluation: a good output must apply the retained edits, \emph{avoid} the suppressed edit, and otherwise leave the query image \(B\) untouched. No single similarity score captures these three requirements at once. A method that copies the full demonstration scores well on \emph{``did the edits appear?''}, but fails on suppression. A method that barely changes \(B\) scores well on preservation but fails to apply anything. We therefore evaluate each output with a structured, \emph{per-edit} protocol rather than a single holistic judgment.

The main paper reports five metrics: three VLM-judge metrics (\emph{Retained Accuracy}, \emph{Preservation}, and \emph{Edit Recall}) and two CLIP-based metrics (\emph{CLIP-Dir Image} and \emph{CLIP-Dir Retained}); we additionally report a \emph{Suppression} score in the ablation (Tab.~\ref{tab:ablation}). Below we give the full protocol behind these numbers: the structured VLM evaluation, the deterministic per-edit mapping, the consistency correction, and the SIA-aware extension of the LoRWeB~\cite{manor2026lorweb} CLIP-directional metric. All metrics are computed for each (sample, method) pair and then aggregated in two stages: (i) the mean within each task triple (setting, data id, retained subset), and (ii) the mean across tasks. This two-step aggregation follows LoRWeB and prevents recipes with many retained-subset choices from dominating the final average.

\subsection{VLM single-image protocol}
\label{supp:metrics:vlm}
Rather than asking the judge for a single overall rating, which tends to be noisy and hard to reproduce, we ask it to reason about \emph{one edit at a time} and commit to a discrete label, which we then map to a number through a fixed table. This keeps the judgment grounded in specific, checkable observations and removes most of the subjectivity from the numeric score.

For each result, we show the judge the four images $[A,A',B,\hat{B}']$ with the retained and suppressed edit lists. The judge follows a three-step evaluation protocol:

\begin{itemize}[leftmargin=*]
    \item \textbf{STEP 1.} Give a brief factual description of \(\hat{B}'\). This anchors the later judgments in what is actually visible, rather than in what the edit lists say \emph{should} be there.
    \item \textbf{STEP 2.} For each atomic edit $e\in\Ecal$, describe the $A\!\to\!A'$ change in the aspect named by $e$, describe the corresponding $B\!\to\!\hat{B}'$ change, and assign a per-edit match label $m(e)\in\{a,b,c\}$. The label records how closely the output reproduces the \emph{specific} demonstrated change, not merely whether the right category of edit occurred:
        \begin{itemize}[nosep]
            \item $(a)$ $B\!\to\!\hat{B}'$ reproduces $A\!\to\!A'$ exactly, including the same specific instance.
            \item $(b)$ the same aspect changes, but with a different specific instance.
            \item $(c)$ the aspect is unchanged.
        \end{itemize}
    The judge also assigns an identity-match label $m_{\mathrm{id}}\in\{a,b,c\}$ indicating whether $\hat{B}'$ preserves the identity, layout, and unrelated content of $B$.
    \item \textbf{STEP 3.} Map each per-edit match label deterministically to a binary detection flag $d(e)\in\{0,1\}$ and an integer score $s(e)\in\{1,\ldots,10\}$ using Tab.~\ref{tab:match_score_map}. The identity-match label is mapped to a 1--10 Preservation score using the same score ranges.
\end{itemize}
The model returns a flat JSON object containing the description, the per-edit fields $(m,d,s)$ for every $e\in\Ecal$, and the identity-match label with its 1--10 Preservation score; the exact prompt and output schema are given in Sec.~\ref{supp:vlm_prompts}. We extract these fields and apply the small consistency correction described below before aggregation.

\begin{table}[h]
\centering
\caption{Deterministic mapping from the per-edit match label in STEP~2 to the binary detection flag $d(e)$ and the 1--10 score $s(e)$ in STEP~3. The same score ranges are used to map the identity-match label to the Preservation score.}
\label{tab:match_score_map}
\small
\begin{tabular}{llcc}
\toprule
Role & Match $m(e)$ & $d(e)$ & $s(e)$ (1--10) \\
\midrule
\textsc{retain}   & $(a)$ & 1 & 9--10 \\
\textsc{retain}   & $(b)$ & 1 & 5--7  \\
\textsc{retain}   & $(c)$ & 0 & 1     \\
\midrule
\textsc{suppress} & $(a)$ & 1 & 1     \\
\textsc{suppress} & $(b)$ & 1 & 4--6  \\
\textsc{suppress} & $(c)$ & 0 & 9--10 \\
\bottomrule
\end{tabular}
\end{table}

\paragraph{Judges.}
We report results with three independent judges. The primary judge is Qwen3-VL-32B-Instruct~\cite{Qwen3-VL}. As robustness checks, we additionally report scores from Gemma~4 (gemma-4-31B-it)~\cite{gemmateam2026gemma4}, which is also shown in the main paper, and from Gemma~3~\cite{gemmateam2025gemma3}, which is reported only here. All judges use the identical prompt, metrics, and aggregation. They differ in absolute calibration, with Gemma~3 being more lenient overall, but all three rank \SIA{} first on Retained Accuracy and Edit Recall. Tab.~\ref{tab:quantitative_full} reports the full comparison under all three judges.

\begin{table*}[t]
\caption{Full quantitative comparison with visual analogy baselines under all three VLM judges (extends Tab.~\ref{tab:quantitative} with Gemma 3). Ret. Acc.
measures retained-edit quality, Pres. measures preservation of $B$, and
Recall measures retained-edit recall. CLIP-Dir Image is the image-image
directional metric, while CLIP-Dir Retained measures alignment with the
retained edit direction.}
\label{tab:quantitative_full}
\small
\setlength{\tabcolsep}{3.2pt}
\begin{tabular}{l ccc ccc ccc cc}
\toprule
& \multicolumn{3}{c}{Qwen3-VL-32B} & \multicolumn{3}{c}{Gemma 3}
& \multicolumn{3}{c}{Gemma 4} & \multicolumn{2}{c}{CLIP} \\
\cmidrule(lr){2-4}\cmidrule(lr){5-7}\cmidrule(lr){8-10}\cmidrule(lr){11-12}
Method & Ret. Acc.$\uparrow$ & Pres.$\uparrow$ & Recall$\uparrow$
       & Ret. Acc.$\uparrow$ & Pres.$\uparrow$ & Recall$\uparrow$
       & Ret. Acc.$\uparrow$ & Pres.$\uparrow$ & Recall$\uparrow$
       & Dir. Image$\uparrow$ & Dir. Ret.$\uparrow$ \\
\midrule
RelationAdapter~\cite{gong2025relationadapter}
  & 6.74 & 9.93 & 0.782 & 8.09 & 9.48 & 0.941 & 6.14 & 9.50 & 0.706 & 0.386 & 0.144 \\
VisualCloze~\cite{li2025visualcloze}
  & 7.24 & 8.62 & 0.856 & 8.14 & 9.28 & 0.947 & 6.35 & 7.53 & 0.743 & 0.464 & 0.152 \\
Edit Transfer~\cite{chen2025edittransfer}
  & 2.31 & \textbf{9.96} & 0.176 & 5.75 & \textbf{9.58} & 0.623 & 2.35 & 9.83 & 0.181 & 0.102 & 0.023 \\
LoRWeB~\cite{manor2026lorweb}
  & 6.59 & 9.83 & 0.736 & 8.12 & 9.51 & 0.926 & 5.27 & 9.72 & 0.567 & 0.402 & 0.155 \\
\midrule
\textbf{Ours}
  & \textbf{8.01} & 9.75 & \textbf{0.892} & \textbf{8.65} & 9.57 & \textbf{0.952}
  & \textbf{9.31} & \textbf{9.89} & \textbf{0.965} & \textbf{0.543} & \textbf{0.171} \\
\bottomrule
\end{tabular}
\end{table*}

\paragraph{Consistency correction.}
The 1--10 score and the binary detection flag are both derived from the same match label. However, VLM responses occasionally break this coupling, for example by assigning a high numeric score while also marking the edit as undetected. To keep Retained Accuracy consistent with the detection flags, we cap it at $2$ when no retained edit is detected and at $5$ when fewer than half of the retained edits are detected. No other corrections are applied to the metrics reported in the main paper.

\subsection{VLM metric definitions}
\label{supp:metrics:single}

Let $s(e)\in\{1,\ldots,10\}$ denote the consistency-corrected per-edit score, and let $d(e)\in\{0,1\}$ denote the detection flag. Recall that \(\Rcal\) is the retained-edit set and \(\Scal=\Ecal\setminus\Rcal\) is the suppressed-edit set.

\paragraph{Retained Accuracy~$\uparrow$.}
Retained Accuracy is the average per-edit score over the retained edits:
\[
    \mathrm{RetAcc} \;=\; \frac{1}{|\Rcal|}\sum_{e\in\Rcal} s(e).
\]

\paragraph{Suppression~$\uparrow$.}
Suppression is the analogous average over the \emph{suppressed} edits, and measures how well the output leaves out the edits the user chose to exclude:
\[
    \mathrm{Supp} \;=\; \frac{1}{|\Scal|}\sum_{e\in\Scal} s(e).
\]
Because the suppress role uses the inverted score ranges of Tab.~\ref{tab:match_score_map}, a high Suppression score means the suppressed edits are correctly \emph{absent} from \(\hat{B}'\). This metric targets the core requirement of selective transfer, which Preservation alone does not capture: a method that leaves \(B\) almost unchanged scores well on Preservation yet may still fail to remove a suppressed edit that the demonstration would otherwise introduce. Suppression is defined only for partial-transfer samples, where \(\Scal\neq\emptyset\).

\paragraph{Preservation~$\uparrow$.}
Preservation is the 1--10 score obtained from the STEP-2 identity-match label, using the \textsc{retain} score ranges of Tab.~\ref{tab:match_score_map}: $(a)\!\Rightarrow\!9\text{--}10$, $(b)\!\Rightarrow\!5\text{--}7$, and $(c)\!\Rightarrow\!1$. It reflects whether identity, layout, and unrelated content of \(B\) are left intact. A no-op method that returns $\hat{B}'\!\approx\!B$ will naturally score close to $10$ on this metric, which is why Preservation must always be read together with Retained Accuracy and Suppression rather than on its own.

\paragraph{Edit Recall~$\uparrow$.}
Edit Recall is the fraction of retained edits detected by the VLM:
\[
    \mathrm{EditRecall}
    =
    \frac{1}{|\Rcal|}
    \sum_{e\in\Rcal}
    \mathbf{1}\!\left[d(e)=1\right].
\]
The indicator is $1$ when edit $e$ is detected and $0$ otherwise. Unlike Retained Accuracy, Recall only asks \emph{whether} each wanted edit appeared at all, not whether it matches the demonstrated form. A method can therefore reach high Recall while still scoring lower on Retained Accuracy.

\subsection{CLIP directional metrics}
\label{supp:metrics:clip}

The VLM metrics above are semantic and judge-dependent. We complement them with two lightweight, judge-independent CLIP directional metrics. The original CLIP-directional metric used by LoRWeB~\cite{manor2026lorweb} assumes that $\hat{B}'$ should reproduce the \emph{full} transformation from $A$ to $A'$. In selective image analogy, this assumption only holds for full-transfer samples. When the user suppresses an edit, $\hat{B}'$ should intentionally \emph{disagree} with the full $A\!\to\!A'$ transformation along the suppressed-edit axis. As a result, the original metric can incorrectly penalize a correct selective output on partial-transfer samples. To address this, we also report a text-anchored directional metric built from the per-edit captions stored with each sample, following the text-driven CLIP-direction recipe popularized by StyleCLIP~\cite{patashnik2021styleclip}.

\paragraph{Notation.}
Let $\phi_I$ and $\phi_T$ denote the CLIP image and text encoders (ViT-L/14, OpenAI weights). For a single (sample, method) tuple, define the image-space directions
\[
    \Delta I \;=\; \phi_I(\hat{B}') - \phi_I(B),
    \qquad
    \Delta I_A \;=\; \phi_I(A') - \phi_I(A),
\]
and the retained-text direction
\[
    \Delta T_R \;=\; \phi_T(t_R) - \phi_T(t_{\mathrm{src}}),
\]
where
\[
    t_{\mathrm{src}} = \text{``a photo''},
    \quad
    t_R = \text{``a photo with }\langle\text{retained desc.}\rangle\text{''}
\]
is built from the per-edit captions of the retained set $\Rcal$.

\paragraph{CLIP-Dir Image~$\uparrow$.}
This is the original LoRWeB image-image directional metric, which checks whether the output moves through image space in the same direction as the demonstrated edit:
\[
    \mathrm{CLIPDirImage} \;=\; \cos\;\!\bigl(\Delta I_A,\,\Delta I\bigr).
\]
We compute it only on full-transfer samples ($\Rcal=\Ecal$). It is reported as NaN on partial-transfer rows because $\Delta I_A$ also includes the suppressed edits in those cases and would penalize a correct selective output.

\paragraph{CLIP-Dir Retained~$\uparrow$.}
This is our SIA-aware text-anchored directional metric:
\[
    \mathrm{CLIPDirRetained} \;=\; \cos\;\!\bigl(\Delta I,\,\Delta T_R\bigr).
\]
It replaces the demonstrated image-space direction with the direction described by the retained-edit \emph{text}, so it is defined for every sample, including partial transfers, and measures whether $\hat{B}'$ moves toward the retained-edit description regardless of what the suppressed edits do. It therefore serves as the selective-transfer counterpart of \(\mathrm{CLIPDirImage}\) when no full-transfer ground truth is available.

Both CLIP metrics are judge-independent and are computed once for the full benchmark.

\subsection{VLM Evaluation Prompt}
\label{supp:vlm_prompts}
For reproducibility, we include the evaluation prompt. It restates the three-step protocol of Sec.~\ref{supp:metrics:vlm}. The box fixes the exact output schema, which we parse to obtain the per-edit fields used by the metrics in Sec.~\ref{supp:metrics:single}. The VLM receives the four images $[A,A',B,\hat{B}']$ together with the retained- and suppressed-edit lists.

\begin{tcolorbox}[
    breakable,
    enhanced,
    colback=gray!8,
    colframe=gray!50!black,
    boxrule=0.4pt,
    arc=1.5mm,
    left=3pt, right=3pt, top=3pt, bottom=3pt,
    title=\textbf{VLM Evaluation Prompt},
    fonttitle=\small,
    colbacktitle=gray!25,
    coltitle=black,
    width=\columnwidth,
    fontupper=\small
]
You are evaluating a selective image analogy result.
You are given: \\
(1) Image $A$: the original demonstration image. \\
(2) Image $A'$: the edited version of Image $A$. \\
(3) Image $B$: the query image to edit. \\
(4) Output $\hat{B}'$: the generated result. \\
(5) Retained edits: edits that should appear in the output. \\
(6) Suppressed edits: edits that should not appear in the output. \\
\textbf{STEP 1.} Describe the output $\hat{B}'$ briefly. \\
\textbf{STEP 2.} For each atomic edit $e_i$ in the demonstrated set, describe the $A\!\to\!A'$ change in the aspect named by $e_i$, describe the corresponding $B\!\to\!\hat{B}'$ change, and commit to a match label in $\{a,b,c\}$:
\begin{itemize}[nosep]
    \item $(a)$ same aspect, same specific instance.
    \item $(b)$ same aspect, different specific instance.
    \item $(c)$ aspect unchanged.
\end{itemize}
\textbf{STEP 3.} Use the match label to deterministically commit to a detection flag and a 1--10 score using the following role-dependent mapping. For a \textsc{retain} edit: 
\begin{itemize}[nosep]
    \item $(a)$ gives detected $=1$ and score $9$--$10$; 
    \item $(b)$ gives detected $=1$ and score $5$--$7$;
    \item $(c)$ gives detected $=0$ and score $1$. 
\end{itemize}
For a \textsc{suppress} edit: 
\begin{itemize}
    \item $(a)$ gives detected $=1$ and score $1$; 
    \item $(b)$ gives detected $=1$ and score $4$--$6$; 
    \item $(c)$ gives detected $=0$ and score $9$--$10$. 
\end{itemize}
Use the same score ranges to map the identity-match label to the preservation score. Return your answer as a flat JSON object: \\
\ttfamily\footnotesize
\noindent \{\\
\hspace*{1em}"step1": "<brief description of B'>",\\
\hspace*{1em}"per\_edit": [\\
\hspace*{2em}\{\\
\hspace*{3em}"edit": "<edit phrase>",\\
\hspace*{3em}"role": "retain" | "suppress",\\
\hspace*{3em}"a\_to\_a\_prime": "<sentence>",\\
\hspace*{3em}"b\_to\_b\_hat": "<sentence>",\\
\hspace*{3em}"match": "a" | "b" | "c",\\
\hspace*{3em}"detected": true | false,\\
\hspace*{3em}"score": <integer 1..10>\\
\hspace*{2em}\},\\
\hspace*{2em}...\\
\hspace*{1em}],\\
\hspace*{1em}"identity\_match": "a" | "b" | "c",\\
\hspace*{1em}"preservation": <integer 1..10>
\}\\
\end{tcolorbox}

\section{Additional Results}
\label{supp:additional_results}
\begin{figure*}[p]
    \centering
    \includegraphics{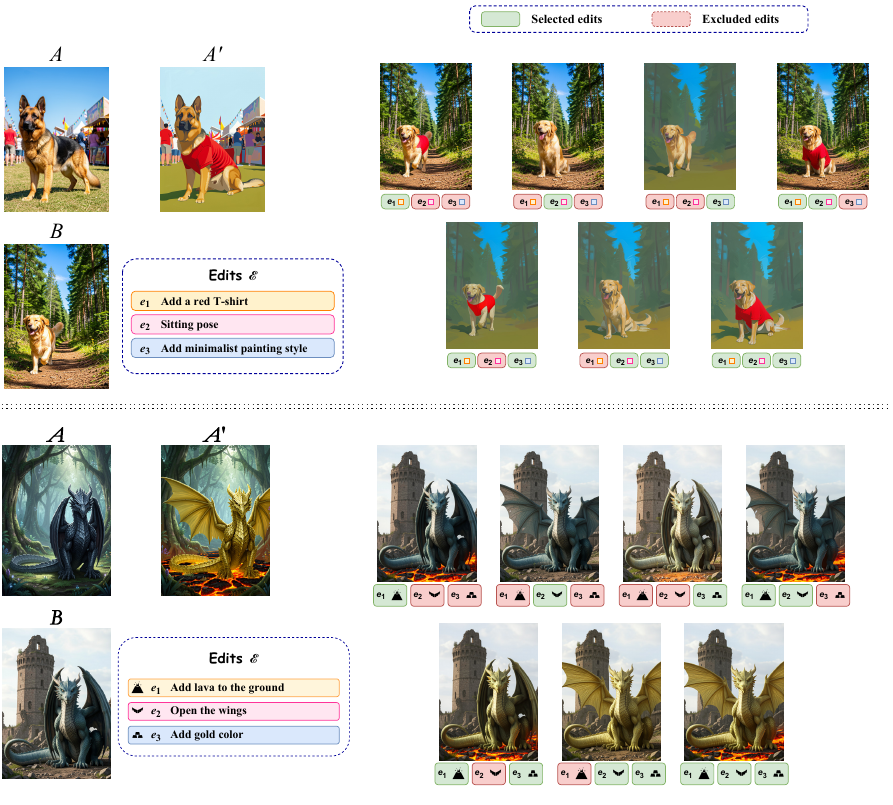}
    \caption{\textbf{All edit combinations.} For each demonstration pair $(A, A')$ and query image $B$, we show the output of \SIA{} for every subset of the demonstrated edits, from a single retained edit to all of them. The chips under each output mark the selected (green) and excluded (red) edits.
    }
    \label{fig:supp_all_combinations}
\end{figure*}

\begin{figure*}[p]
    \centering
    \includegraphics{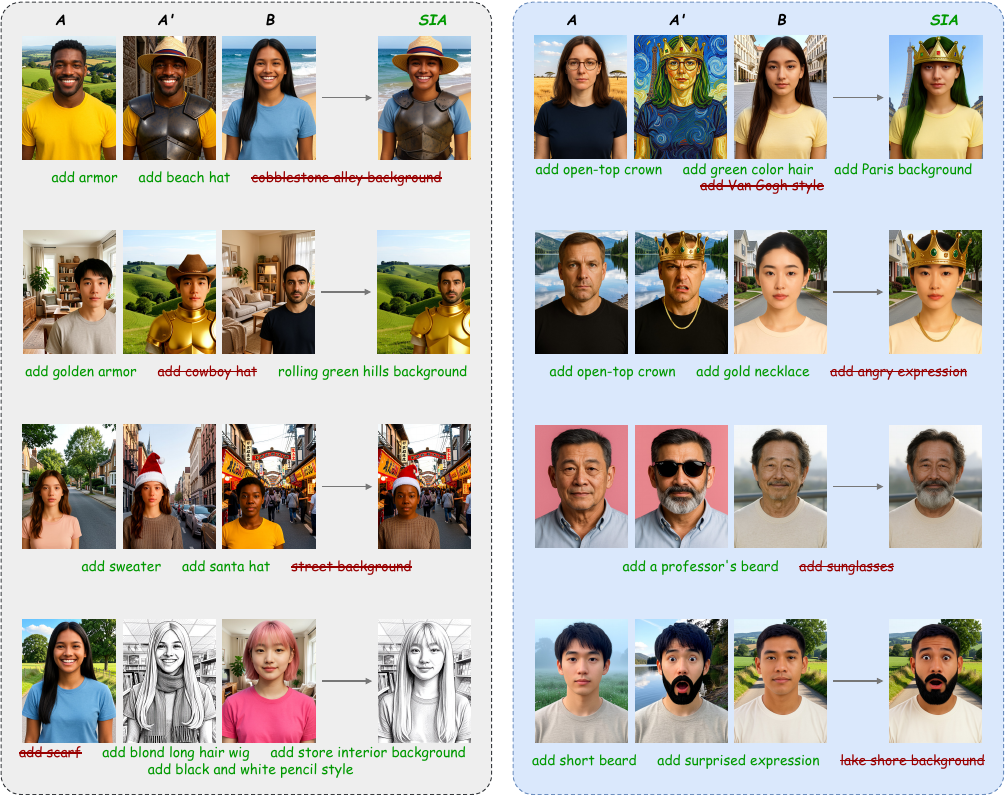}
    \caption{\textbf{Additional Results on portraits with multi-edit demonstrations.} Each row shows SIA transferring a selected subset of the edits in a demonstration pair $(A, A')$ to a query image $B$. Green edits are retained and struck-through red edits are suppressed. The examples cover accessories, hair, expressions, backgrounds, and styles on portraits, from demonstrations with two to four edits.
    }
    \label{fig:supp_portraits_multiedit}
\end{figure*}

\begin{figure*}[p]
    \centering
    \includegraphics{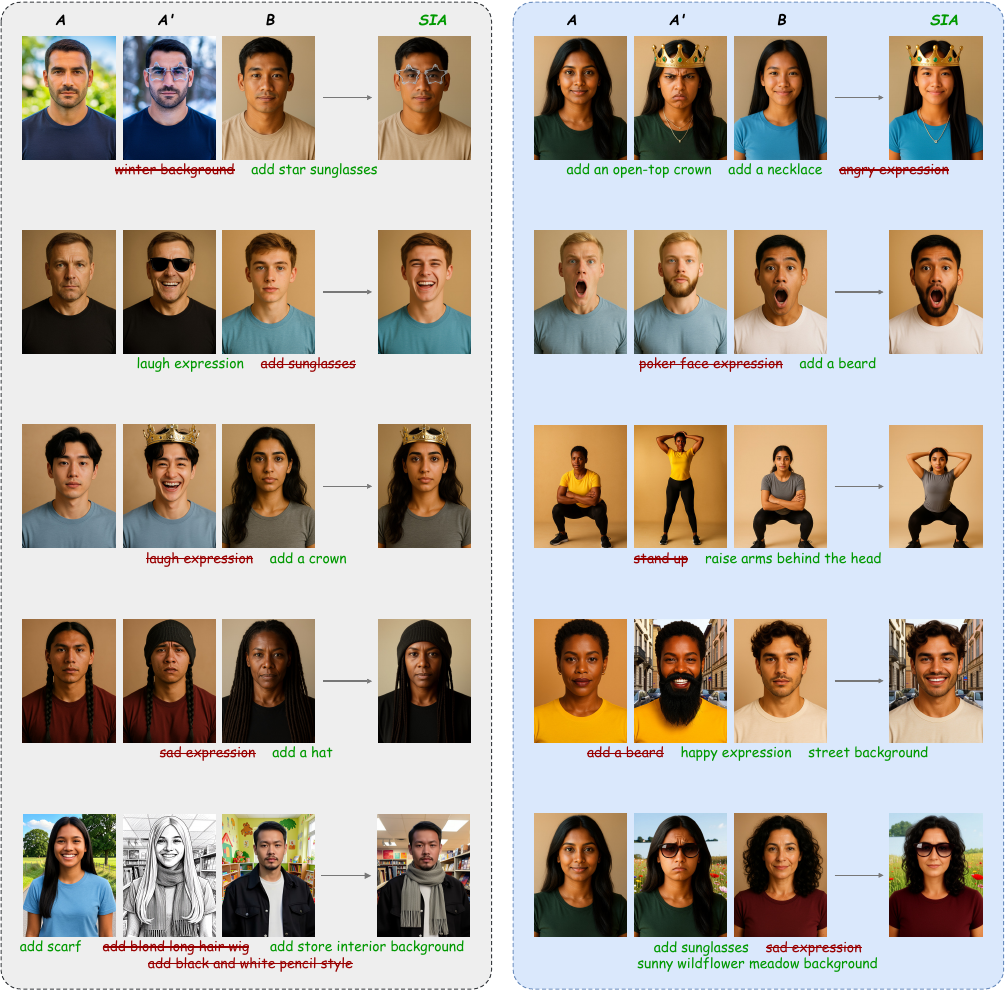}
    \caption{\textbf{Additional Results on portraits.} Each row shows SIA transferring a selected subset of the edits in a demonstration pair $(A, A')$ to a query image $B$. Green edits are retained and struck-through red edits are suppressed. The examples cover expressions, accessories, poses, and backgrounds, mostly from two-edit demonstrations.
    }
    \label{fig:supp_portraits}
\end{figure*}

\begin{figure*}[p]
    \centering
    \includegraphics{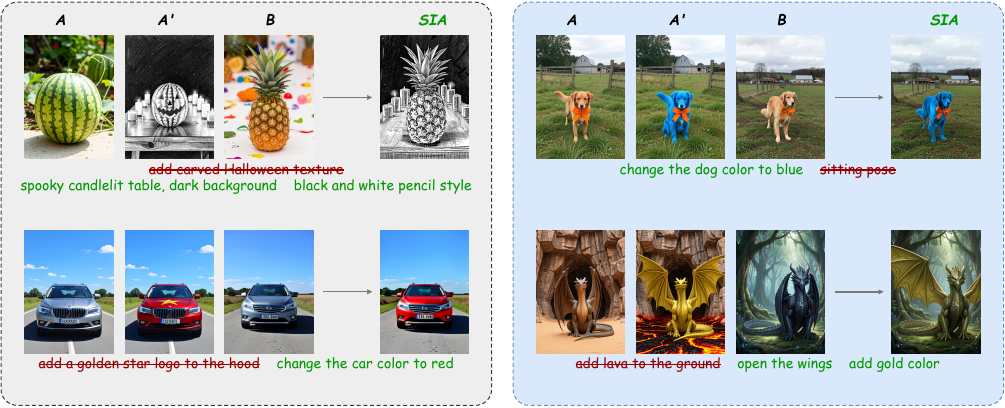}
    \caption{\textbf{Additional Results on objects, scenes, and animals.} Each row shows SIA transferring a selected subset of the edits in a demonstration pair $(A, A')$ to a query image $B$. Green edits are retained and struck-through red edits are suppressed. The examples cover candlelit-scene and pencil-sketch edits on fruit, a color change on a car, a color change on a dog, and wing and gold-color edits on a dragon.
    }
    \label{fig:supp_objects_animals}
\end{figure*}

Figs.~\ref{fig:supp_all_combinations}--\ref{fig:supp_objects_animals} show further results of \SIA{} with the \flux-Klein{} model used throughout the paper. Fig.~\ref{fig:supp_all_combinations} shows every combination of edits for two demonstrations, while the remaining figures show one selected subset per row, for portraits, objects, scenes, and animals. Sec.~\ref{sec:qwen_image} reports results of a second backbone, Qwen-Image-2.1, in Figs.~\ref{fig:qwen_results}--\ref{fig:qwen_comb3}.

\clearpage

\section{SIA on Qwen-Image-2.1}
\label{sec:qwen_image}
\begin{figure*}[p]
    \centering
    \includegraphics{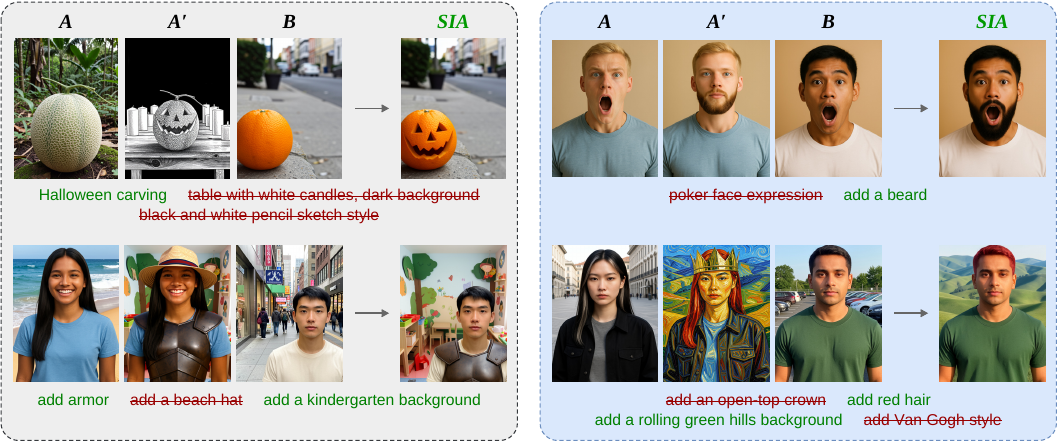}
    \caption{\textbf{Additional results with Qwen-Image-2.1.} Each row shows SIA transferring a selected subset of the edits in a demonstration pair $(A, A')$ to a query image $B$. Green edits are retained and struck-through red edits are suppressed. The examples cover a carved texture on a fruit, armor with a new background, a beard, and hair color with a background. The beard row uses the same demonstration pair and query image as a row of Fig.~\ref{fig:supp_portraits}, for a side-by-side view of the two backbones. These are curated examples (Sec.~\ref{sec:qwen_image}), and no baseline or user-study comparison exists for this backbone.}
    \label{fig:qwen_results}
\end{figure*}

\begin{figure*}[p]
    \centering
    \includegraphics{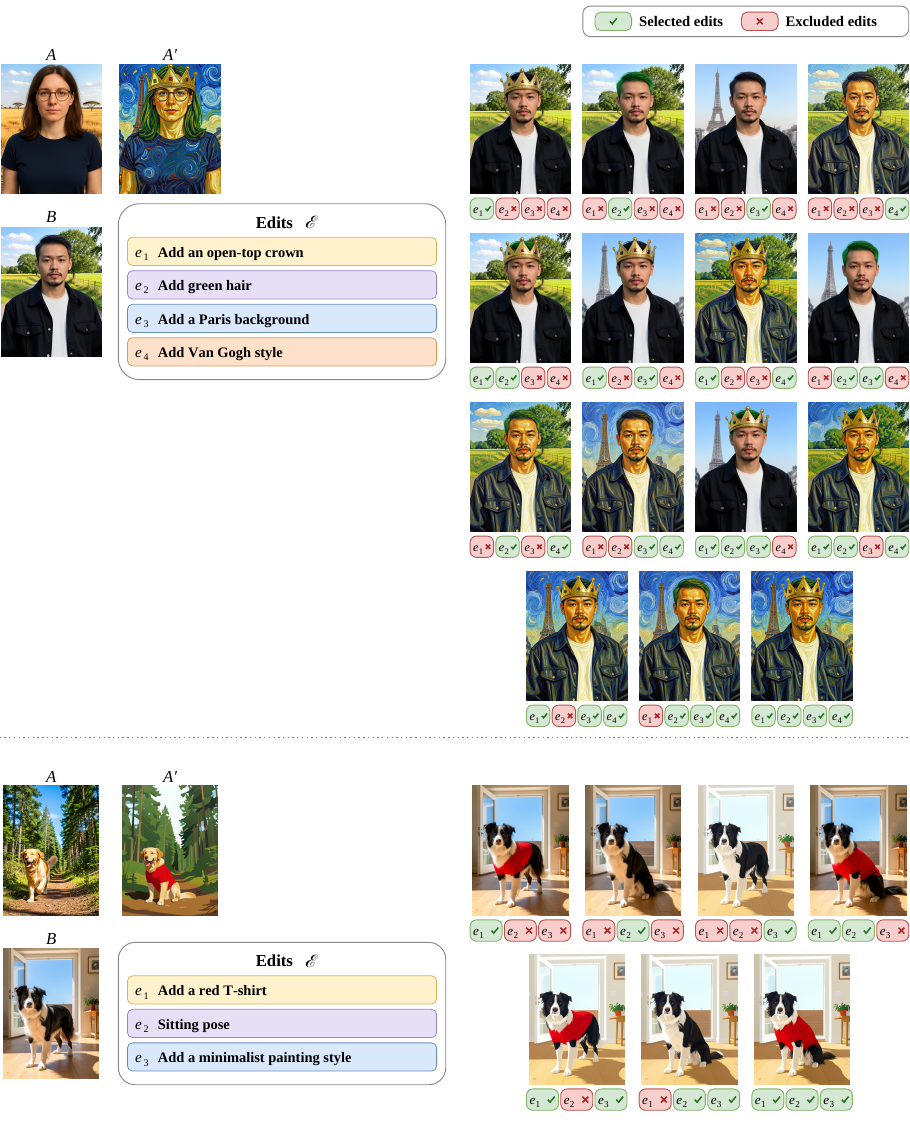}
    \caption{\textbf{All edit combinations with Qwen-Image-2.1.} For each demonstration pair $(A, A')$ and query image $B$, we show the output of \SIA{} for every subset of the demonstrated edits: a crown, green hair, a Paris background and a Van Gogh style on a portrait (four edits), and a red T-shirt, a sitting pose and a painting style on a dog. The chips under each output mark the selected (green, tick) and excluded (red, cross) edits. All outputs are curated examples from the Qwen-Image-2.1 model of Sec.~\ref{sec:qwen_image}, not from \flux-Klein{}, and no baseline or user-study comparison exists for this backbone.}
    \label{fig:qwen_comb1}
\end{figure*}

\begin{figure*}[p]
    \centering
    \includegraphics{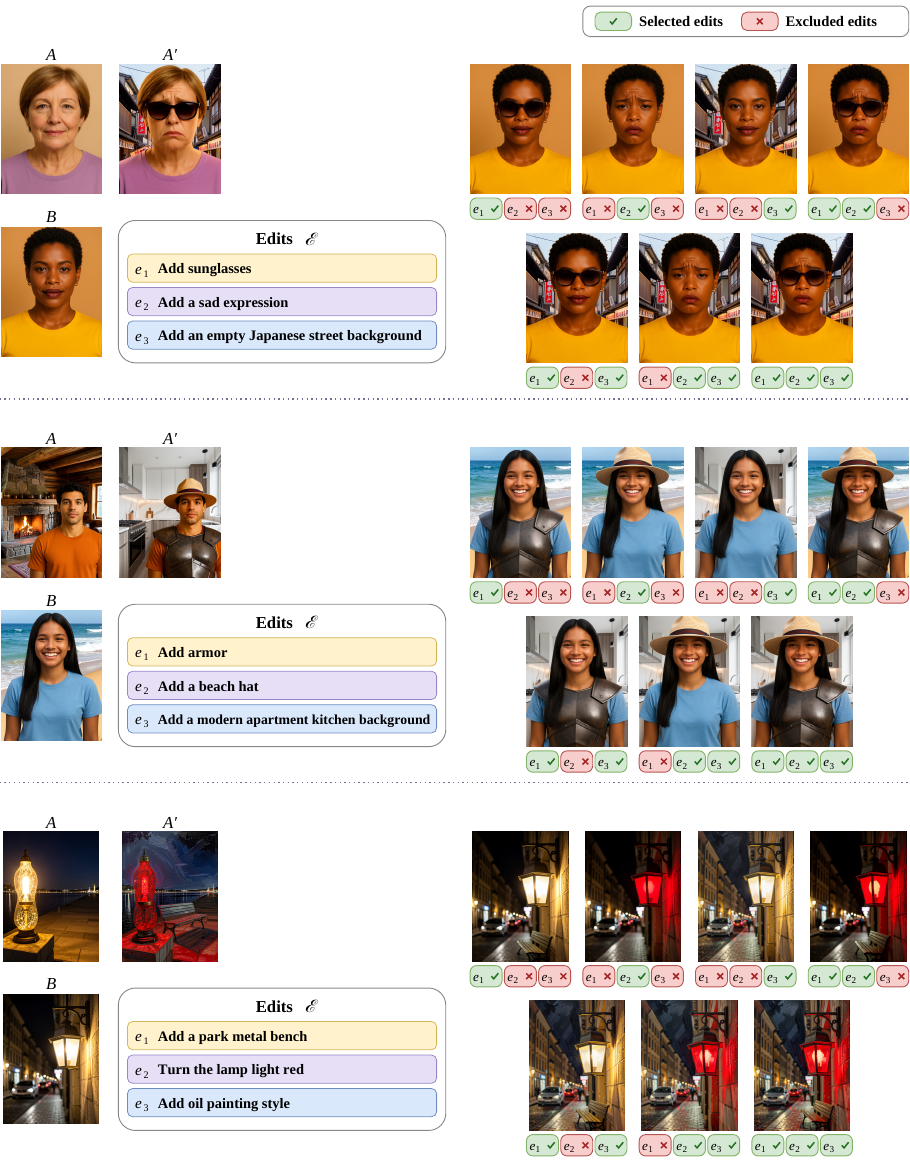}
    \caption{\textbf{All edit combinations with Qwen-Image-2.1 (continued).} More demonstrations in the format of Fig.~\ref{fig:qwen_comb1}: sunglasses, a sad expression and a street background; armor, a beach hat and a kitchen background; and a bench, a red lamp light and an oil-painting style on a street scene.}
    \label{fig:qwen_comb2}
\end{figure*}

\begin{figure*}[p]
    \centering
    \includegraphics{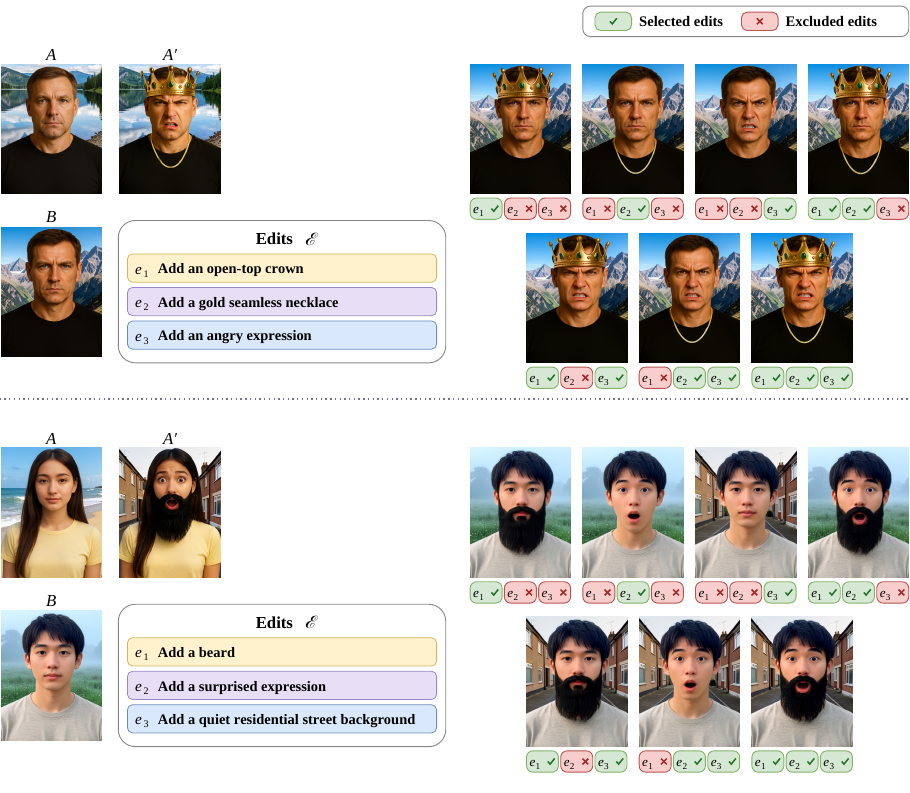}
    \caption{\textbf{All edit combinations with Qwen-Image-2.1 (continued).} A crown, a gold necklace and an angry expression; and a beard, a surprised expression and a street background. These are curated examples from the Qwen-Image-2.1 model (Sec.~\ref{sec:qwen_image}).}
    \label{fig:qwen_comb3}
\end{figure*}

All experiments in the main paper, including the quantitative tables, the comparisons with baselines, the ablations and the user studies, use \flux-Klein{}. In addition, we trained \SIA{} on a second, more recent backbone, Qwen-Image-2.1, from the Qwen-Image family~\cite{qwen2025imageedit}. This suggests that the method is not tied to one backbone: the same two stages and the same training data carry over. In our informal inspection the Qwen-Image-2.1 outputs also look visually better than the \flux-Klein{} ones, possibly because the backbone is more recent; we did not measure this. The Qwen-Image-2.1 model uses \AnaLoRA{} and \ExLoRA{} without \CSV{}, and we did not repeat the baseline comparisons, the ablations or the user studies for it, so every claim in the main paper refers to \flux-Klein{} only. Figs.~\ref{fig:qwen_results}--\ref{fig:qwen_comb3} show curated results of this model.

\paragraph{Training.}
We follow the two-stage recipe of Sec.~\ref{supp:impl} and train on the same training split of our multi-edit analogy dataset (Sec.~\ref{supp:dataset}). Both stages use Qwen-Image-2.1 at 512\,px, LoRA rank $r{=}16$, batch size 16 and seed 0. The learning rates were chosen with a short sweep of \AnaLoRA{} runs, so they, and the \ExLoRA{} batch size, differ from the \flux-Klein{} values. \AnaLoRA{} is trained with learning rate $2{\times}10^{-4}$, with the prompt dropped for 15\% of the samples, and we use the checkpoint after 2K steps. \ExLoRA{} is initialized from that checkpoint and trained with learning rate $2{\times}10^{-4}$, while the jointly updated \AnaLoRA{} weights use $5{\times}10^{-5}$ and an anchoring weight of $0.1$, as for \flux-Klein{}. \ExLoRA{} is trained for about $6.5$K steps and we use the checkpoint after 5K steps. This checkpoint was selected on half of the test set, where the checkpoints between 3K and 6K steps are close (average $9.55$--$9.62$), so the scores below are slightly optimistic. To apply the suppression mask to the prompt tokens of the excluded edits, we map the tokens of each edit phrase to the rows of the prompt embedding and to the joint text--image token sequence consumed by the transformer. At inference we use 40 denoising steps without classifier-free guidance.

\paragraph{Prompt order at inference.}
For the figures in this section, the retained edits are listed first in the prompt and the excluded edits last. This is a change of the prompt only and needs no retraining. On a quarter of the benchmark (352 items from 60 cases) it raises the judge average by $0.20$ over the default order (the output is better for 34 items, worse for 11 and unchanged for 307; sign test $p{=}0.0008$). All of the gain is on edit subsets that the training data does not cover.

\paragraph{Evaluation.}
We evaluate on the 239 test cases, scoring every non-empty proper subset of the edits ($1{,}406$ items) with Qwen3-VL-32B-Instruct, using one image and one seed per item. The judge scores Retained Accuracy, Suppression and Preservation on a 0--10 scale. Unlike Sec.~\ref{supp:metrics}, this protocol has no consistency correction and uses a plain mean, so the numbers are not comparable with the main tables. Tab.~\ref{tab:qwen} compares \ExLoRA{} with deleting the excluded edits from the \AnaLoRA{} prompt: prompt deletion lowers Suppression from $9.94$ to $7.26$. The judge is lenient, however: returning $B$ unchanged scores $10.00$ on Suppression and Preservation but only $1.09$ on Retained Accuracy, so a high Suppression score alone is not evidence of quality. Adding \CSV{} on top of the trained \ExLoRA{} gave no measurable gain, so we use \ExLoRA{} without \CSV{} on Qwen-Image-2.1.

\begin{table}[!htb]
\caption{\textbf{SIA on Qwen-Image-2.1.} Mean judge scores ($0$--$10$, higher is better) from Qwen3-VL-32B-Instruct on a $704$-item subset of the test set. Both variants run without \CSV{}. Prompt deletion drops the excluded edits from the \AnaLoRA{} prompt. Avg.\ is the mean of the three scores. The protocol differs from that of the main tables.}
\label{tab:qwen}
\centering
\small
\setlength{\tabcolsep}{4pt}
\begin{tabular}{>{\raggedright\arraybackslash}p{3cm}cccc}
\toprule
Variant & Ret.\ Acc. & Supp. & Pres. & Avg. \\
\midrule
\AnaLoRA{} + \ExLoRA{} & 9.12 & 9.94 & 9.80 & 9.62 \\
\addlinespace
\AnaLoRA{} only, prompt deletion & 9.16 & 7.26 & 8.87 & 8.43 \\
\bottomrule
\end{tabular}
\end{table}

\paragraph{What the figures show, and what they do not.}
The cases in Figs.~\ref{fig:qwen_results}--\ref{fig:qwen_comb3} were selected from the 60 benchmark cases rendered with that prompt order. We kept cases without a visible leak of an excluded edit, without a mismatch between a retained edit and the one in $A'$, and without artifacts. They show what the model does well and are not a random sample. We also saw failures that the judge does not penalize strongly. The attribute that is added to a query image tends to take a generic form for that person, so an accessory matches the one in $A'$ closely only when $A'$ already resembles that generic form. A style edit can tint hair or clothes, a background edit is subtle when $B$ already shows a similar scene, and a style or pencil edit is sometimes only half applied when the background edit is dropped. The four-edit case of Fig.~\ref{fig:qwen_comb1} has weak tiles as well: with the crown, the green hair and the Paris background retained, and no style, the subject is reframed and smaller than in $B$, and in several tiles the crown hides most of the green hair. Because the benchmark covers proper subsets only, the all-edits outputs in Figs.~\ref{fig:qwen_comb1}--\ref{fig:qwen_comb3} were generated separately with the same weights (\AnaLoRA{} at 2K and \ExLoRA{} at 5K steps), without suppression, with seed 0, 40 steps and 512\,px.

\end{document}